\documentclass[11pt]{article}

\usepackage[final]{acl}
\usepackage{times}
\usepackage{latexsym}
\usepackage{amssymb}
\usepackage{amsmath}
\usepackage{pifont}
\usepackage{xcolor}
\usepackage{booktabs}
\usepackage{hyperref}
\usepackage{url}
\usepackage{booktabs}
\usepackage{multirow}

\usepackage[T1]{fontenc}

\usepackage[utf8]{inputenc}
\usepackage{CJKutf8}

\usepackage[nopatch=footnote]{microtype}

\usepackage{inconsolata}
\usepackage{fvextra}

\usepackage{graphicx}
\usepackage{rotating}
\usepackage{float}
\usepackage{setspace}

\newcommand{\cmark}{\textcolor{green!60!black}{\ding{51}}}
\newcommand{\xmark}{\textcolor{red!70!black}{\ding{55}}}

\title{SHAM-Bench: A Span-level Hallucination Benchmark Across \\ Multiple Tasks}

\author{Sicong Huang \and Ian Lane \\ \\
University of California, Santa Cruz \\
\texttt{shaun213@ucsc.edu}, \texttt{ialane@ucsc.edu}}

\begin{document}

\maketitle
\begin{abstract}
Span-level hallucination detection pinpoints unsupported content in LLM outputs, enabling actionable feedback and targeted mitigation.
Existing evaluations for span-level hallucination detection methods are fragmented across tasks, annotation formats, and metrics, and most span-level datasets cover only a single NLG domain, making it difficult to assess whether a detection method actually generalizes across tasks and domains.
In this paper, we introduce \textbf{SHAM-Bench} (\textbf{S}pan-level \textbf{H}allucination benchmark \textbf{A}cross \textbf{M}ultiple tasks), a unified benchmark for span-level hallucination detection spanning six NLG tasks, aggregated from existing human-annotated datasets totaling 6,461 examples under a common schema and evaluation protocol.
Using SHAM-Bench, we evaluate existing hallucination detection methods and find that fine-tuned detectors often perform well in-domain but degrade under distribuiton shift, whereas prompt-based methdos with reasoning are more robust across tasks.
Our analysis further shows that subtle unsupported content remains difficult to detect and span-level evaluation is still challenged by ambiguous span boundaries and noisy labels.
We release the dataset, evaluation code, and a public leaderboard to support future research on broadly generalizable hallucination detection.

\end{abstract}

\section{Introduction}
\label{sec:intro}
While Large Language Models (LLMs) are extremely effective at generating fluent text, they often produce hallucinations, generating content that is \textit{factually incorrect with respect to world knowledge or is unsupported by the available evidence} \citep{huang2025survey, zhang2025sirens}.
Detecting hallucinations in LLM outputs is important for applications ranging from dialog assistants \citep{li-etal-2023-halueval} and decision-support systems \citep{Singhaletal2023, Brynjolfsson2023} to complex agentic workflows \citep{lin2025llmbasedagentssufferhallucinations}.
Early work in hallucination detection focused on coarse-grained, response-level decisions \citep{wang-etal-2020-asking, fabbri-etal-2021-summeval, adlakha-etal-2024-evaluating, healy2026internal}.
While useful for screening, response-level detection does not identify \textit{where} hallucination errors occur, limiting its value for actionable feedback.


Span-level hallucination detection addresses this limitation by localizing the specific words or phrases in a generation that are unsupported or factually incorrect \citep{liu-etal-2022-token}.
Rather than rejecting an entire response, a span-level detector can preserve supported content while identifying hallucinated segments.
This is especially valuable for post-editing workflows \citep{tonmoy2024comprehensivesurveyhallucinationmitigation} and for providing fine-grained training signals for hallucination mitigation \citep{wen-etal-2025-policy}.

Despite its practical importance, span-level hallucination detection remains difficult to evaluate systematically.
Most annotated resources provide only sentence- or response-level labels \citep{lin-etal-2022-truthfulqa}, too coarse to test whether methods identify precise unsupported spans.
Datasets with span-level annotations remain scarce and typically focus on a single task \citep{tang-etal-2023-understanding, vazquez-etal-2025-semeval}, limiting how far their conclusions generalize across domains.
Moreover, differences in annotation formats and evaluation metrics further complicate direct comparisons of detectors.

To address these challenges, we introduce \textbf{SHAM-Bench}, the \textbf{S}pan-level \textbf{H}allucination benchmark \textbf{A}cross \textbf{M}ultiple tasks.
SHAM-Bench is a unified benchmark for span-level hallucination detection.
It consolidates nine task partitions aggregated from six human-annotated datasets into a single standardized multitask benchmark.
The benchmark spans six NLG tasks: open-domain QA, context-based QA, data-to-text generation, biographical generation, summarization, and machine translation, totaling 6{,}461 test set examples.
By normalizing annotation formats and using standard evaluation metrics across all datasets and tasks, SHAM-Bench allows a meaningful comparison of hallucination detection methods across a wide variety of tasks.

Using SHAM-Bench, we perform a comprehensive evaluation of seven existing hallucination detection methods and analyze why they succeed or fail across tasks.
Our results show that fine-tuned detectors often perform well in narrow domains but struggle under distribution shift, while prompt-based methods with reasoning are more robust across domains.
Furthermore, our analysis shows that both annotation and modeling challenges exist. Span annotations can suffer from ambiguous span boundaries and noisy labels, and subtle unsupported content remains difficult to detect reliably even for the most capable model.

Taken together, our findings suggest that advancing span-level hallucination detection requires both accurrate and reliable annotations across tasks to support evaluation and more sensitive detection methods to identify subtle errors.
We hope SHAM-Bench will serve as a valuable resource for the community, allowing practitioners 
to understand the pros and cons of different hallucination detection methods, and to advance 
the state-of-the-art in span-level hallucination detection.
Further, we hope that future work would build on SHAM-Bench to incorporate more 
tasks and datasets capturing the fuller spectrum of NLG tasks to support more generalizable hallucination detection.


\section{Related Works}
\label{sec:related}

\subsection{Detecting Hallucinations in LLMs}

Most early approaches to detect hallucinations in LLM outputs aim to provide response level judgements. 
Starting with a reference and a response, BERTScore \citep{Zhang*2020BERTScore:} judges a response by scoring the embedding similarity between a reference and the response while BARTScore \citep{yuan2021bartscore} uses the log-likelihood of the response directly as the score.
Taking a reference-free approach, SelfCheckGPT \citep{manakul-etal-2023-selfcheckgpt} estimates factuality by sampling multiple responses and measuring their mutual inconsistency.

Later approaches aim to provide more fine-grained hallucination judgements. 
For instance, AlignScore \citep{zha-etal-2023-alignscore} and MiniCheck \citep{tang-etal-2024-minicheck} fine-tune models to determine whether a sentence aligns or is supported by a given context.
RAGTruth \citep{niu-etal-2024-ragtruth} and FAVA \citep{mishra2024-fava} use span-annotated data to fine-tune models to achieve span-level hallucination detection.
QAGS \citep{wang-etal-2020-asking} and FactScore \citep{min-etal-2023-factscore} operate at the fact level by either asking and answering factual questions about the response or by decomposing the response into individual facts and verifying each one.


Another class of methods leverage model-internal signals. 
LLM-Check \citep{sriramanan2024llmcheck} uses internal attention maps, hidden activations, and output probabilities to detect hallucinations within a single response, while \citet{ogasa-arase-2025-hallucinated} exploits attention irregularities across informative layers to localize hallucinated spans.
Leveraging uncertainty, \citet{Farquhar2024SemanticEntropy} introduces semantic entropy as a hallucination signal, showing hallucinated claims exhibit higher entropy. 


\subsection{Evaluating and Benchmarking Hallucination Detection Methods}
Most prior benchmarks for hallucination detection are limited in one of two ways: they either provide only coarse-grained annotations at the response or sentence level, or they provide fine-grained span annotations but only for a single task.

TruthfulQA \citep{lin-etal-2022-truthfulqa} evaluate response-level hallucinations on QA task and find that larger models are more prone to misconceptions.
AggreFact \citep{tang-etal-2023-understanding} aggregates nine summarization datasets for response-level evaluation, finding that no single metric performs best across the individual datasets and HADES \citep{liu-etal-2022-token} gathers token-level annotations for free-form generation, showing that surrounding context length greatly affects how well hallucinations can be detected.
While these approaches offer valuable insight when evaluating hallucination detection on these tasks, the evaluations in each work are limited to a single task.

HaluEval \citep{li-etal-2023-halueval}, though spans multiple tasks, only perform hallucination detection at the response level.
The same is true for LLM-AggreFact \citep{tang-etal-2024-minicheck} which aggregates 10 datasets for sentence-level factual consistency evaluation.
Mu-SHROOM \citep{vazquez-etal-2025-semeval} and PsiloQA \citep{rykov-etal-2025-models}, despite providing span-level annotations for multiple languages, focus exclusively on the QA task.
To our knowledge, RAGTruth \citep{niu-etal-2024-ragtruth} and \citet{zhou-etal-2021-detecting} are the only two benchmark with human labeled span-label annotations of hallucinations that span multiple tasks.



\section{SHAM-Bench}
\label{sec:benchmark}
\textbf{SHAM-Bench}, the \textbf{S}pan-level \textbf{H}allucination Benchmark \textbf{A}cross \textbf{M}ultiple Tasks is a reference-based benchmark for span-level hallucination detection in LLM outputs. It combines all currently available human-annotated span-labeled hallucination detection datasets, covering nine task-specific test sets from six datasets, into a single unified multitask benchmark. SHAM-Bench spans six NLG tasks: open-domain QA, context-based QA, data-to-text, biographical generation, summarization, and machine translation.
The responses in the benchmark are generated by modern LLMs, allowing SHAM-Bench to reflect the real-world conditions and LLM-generated outputs which hallucination detection systems are likely to encounter.
Additionally, the summarization and QA tasks include multiple dataset partitions, enabling SHAM-Bench to identify generalization failures across datasets that single-dataset benchmarks cannot expose.

\subsection{Benchmark Datasets}
\label{subsec:bench_datasets}

\textbf{FaithBench} \citep{bao-etal-2025-faithbench} is a summarization faithfulness benchmark containing 750 summaries (75 passages $\times$ 10 LLMs), where annotators label spans as \emph{questionable}, \emph{benign}, or \emph{unwanted}.
We convert these into binary hallucinated/non-hallucinated span labels.
\textbf{FAVABench} \citep{mishra2024-fava} contains $\sim$1k model responses to open-domain queries with human annotations that localize factual errors at the span level with a fine-grained error-type taxonomy.
After processing, we retain 266 examples, each augmented with GPT-5-retrieved context for use as reference evidence.
\textbf{HalluEntity} \citep{yeh-etal-2025-halluentity} is an entity-level hallucination dataset built from GPT-3.5-generated biographies.
Following the FActScore pipeline \citep{min-etal-2023-factscore}, atomic facts are verified against Wikipedia and mapped to entity mentions, yielding 157 manually verified examples. Each is paired with GPT-5-retrieved context.
\textbf{Mu-SHROOM} \citep{vazquez-etal-2025-semeval} provides human-annotated span-level hallucination labels for LLM-generated answers to open-domain questions across 14 languages.
We use the English subset of 150 examples, each augmented with GPT-5-retrieved context.
\textbf{RAGTruth} \citep{niu-etal-2024-ragtruth} is a span-annotated RAG benchmark covering question answering, data-to-text, and summarization.
Each of the 2{,}700 test examples pairs a user query and evidence context with an LLM-generated response.
Human annotators highlight spans unsupported by or contradicted by the evidence.
\textbf{Zhou} \citep{zhou-etal-2021-detecting} provides token-level hallucination annotations for summarization and machine translation tasks.
It includes 2{,}000 span-annotated summarization outputs adapted from \citet{maynez-etal-2020-faithfulness}, generated by PtGen, TConvS2S, TransS2S, and BERTS2S \citep{see-etal-2017-get, narayan-etal-2018-dont, vaswani2017attention, rothe-etal-2020-leveraging}, and $\sim$500 MT outputs from TransS2S and mBART \citep{vaswani2017attention, liu-etal-2020-multilingual-denoising}.

\subsection{Dataset Schema, Tasks, and Statistics}
\label{subsec:details}

\begin{table}[t]
  \centering
  \small
  \setlength{\tabcolsep}{4pt}
  \renewcommand{\arraystretch}{1.15}
  \resizebox{\columnwidth}{!}{%
  \begin{tabular}{lrrrrr}
    \toprule
    \textbf{Task \&} & \textbf{\% of} & \textbf{count} & \textbf{response} & \textbf{hallu} & \textbf{hallu} \\
      \textbf{dataset}    & \textbf{total} &     & \textbf{words}        & \textbf{words}     & \textbf{rate}      \\
    \midrule
    \textbf{context\_qa}     & \textbf{13.5\%} & \textbf{875}  & \textbf{131.5} & \textbf{7.1}  & \textbf{0.041} \\
    \hspace{1em}ragtruth     & 13.5\%         & 875            & 131.5           & 7.1           & 0.041 \\
    \midrule
    \textbf{data2text}        & \textbf{13.9\%} & \textbf{900}  & \textbf{182.0} & \textbf{7.2}  & \textbf{0.040} \\
    \hspace{1em}ragtruth     & 13.9\%         & 900            & 182.0           & 7.2           & 0.040 \\
    \midrule
    \textbf{mt}              & \textbf{7.4\%} & \textbf{479}  & \textbf{27.5}  & \textbf{4.0}  & \textbf{0.136} \\
    \hspace{1em}zhou         & 7.4\%          & 479            & 27.5            & 4.0           & 0.136 \\
    \midrule
    \textbf{open\_bio} & \textbf{2.4\%} & \textbf{152}  & \textbf{182.2} & \textbf{41.4} & \textbf{0.227} \\
    \hspace{1em}halluentity  & 2.4\%          & 152            & 182.2           & 41.4          & 0.227 \\
    \midrule
    \textbf{open\_qa}        & \textbf{6.4\%} & \textbf{415}  & \textbf{192.0} & \textbf{32.6} & \textbf{0.210} \\
    \hspace{1em}favabench    & 4.0\%          & 261            & 278.4           & 42.5          & 0.142 \\
    \hspace{1em}mushroom     & 2.4\%          & 154            & 45.6            & 15.7          & 0.326 \\
    \midrule
    \textbf{summ}   & \textbf{56.3\%} & \textbf{3640} & \textbf{62.7}  & \textbf{9.1}  & \textbf{0.294} \\
    \hspace{1em}faithbench   & 11.5\%         & 745            & 101.2           & 14.8          & 0.154 \\
    \hspace{1em}ragtruth     & 13.9\%         & 900            & 124.2           & 3.6           & 0.030 \\
    \hspace{1em}zhou         & 30.9\%         & 1995           & 20.5            & 9.5           & 0.466 \\
    \midrule
    \textbf{total}           & \textbf{100.0\%} & \textbf{6461} & \textbf{97.1}  & \textbf{10.5} & \textbf{0.108} \\
    \bottomrule
  \end{tabular}
  }
 \caption{SHAM-Bench statistics in both task aggregates and dataset breakdowns. Hallucination rate is the proportion of reponse words that are hallucinated. Word counts and hallucination rate are averaged across all examples.}
  \label{tab:hallu_stats}
\end{table}

\paragraph{Standardized schema.}
To reconcile differences in dataset formats, we convert every instance to a unified schema with four main fields, enabling consistent preprocessing and evaluation across tasks:
\vspace{-18pt}
\begin{itemize}
  \setlength\itemsep{-2pt}
  \item \textbf{question}: the original user query or instruction that elicited the response.
  \item \textbf{context}: auxiliary input required to answer the query (e.g., evidence passage, structured record, source document, or target-language text). For tasks that do not provide additional input (e.g. open\_qa, open\_biography), this field contains retrieved context that answers the question, so that the hallucination detection methods that require context can use it.
  \item \textbf{response}: the model-generated response to be evaluated for hallucinations.
  \item \textbf{hallu\_spans}: a list of human-annotated hallucinated spans in the response text.
\end{itemize}

\paragraph{Tasks.}
We assign each instance from the six source datasets (\S\ref{subsec:bench_datasets}) to one of six tasks that cover common NLG use cases:
\vspace{-5pt}
\begin{itemize}
  \setlength\itemsep{-2pt}
  \item \textbf{open\_qa}: open-domain QA where the answering model is given only a question, without any supporting evidence or context.
  \item \textbf{context\_qa}: QA conditioned on an explicit context passage; the model receives both a question and a provided context and is expected to answer using that context.
  \item \textbf{data-to-text}: data-grounded generation where the query targets information in structured data (e.g., a JSON record) and the model produces a natural-language output.
  \item \textbf{open\_biography}: biographical generation where the model is prompted to write a biography of a real person.
  \item \textbf{summarization}: summarization of an input document; the model is given a source text and asked to produce a concise summary.
  \item \textbf{machine\_translation}: translation of a source text from one language to another.
\end{itemize}

We provide a small validation set by sampling five instances for each (dataset, task)-pair, yielding 45 instances in total.
This validation set is used to explore the effectiveness of prompt optimization explored in section \S\ref{subsec:prompt} of this paper.

SHAM-Bench contains 6{,}461 examples spanning six tasks sourced from six datasets (Table~\ref{tab:hallu_stats}). Summarization is the largest category, consisting of 56\% of the total examples, while open\_biography is comparatively small, being only 2.4\% of the total. 
Response lengths differ markedly by task; machine\_translation averages 27.5 words in response while open\_qa averages 192 words. 
Hallucination rate is likewise highly non-uniform across tasks and datasets.
At the task level, average hallucination rates range from \(\sim\)0.04 in data2txt and context\_qa to 0.227 in open\_biography and 0.294 in summarization.
The dataset-level breakdown further reveals within-task differences; summarization contains hallucinated content in 3\% of token on RAGTruth but 46.6\% on Zhou, and open\_qa hallucinates 14.2\% of tokens on FAVABench vs. 32.6\% on Mushroom.
Differences in annotation guidelines and the model used to generate responses likely contribute to the variation in hallucination rates across tasks and datasets.

\subsection{Metrics and Evaluation}
\label{sec:eval}

We use F$_1$ score as the primary evaluation metric.
This choice follows prior span-level hallucination detection work including \citet{mishra2024-fava, niu-etal-2024-ragtruth, zhou-etal-2021-detecting}. 
Moreover, since SHAM-Bench standardizes detector outputs into binary token-level labels, hallucination detection can be naturally evaluated as a binary classification problem. F$_1$ score balances precision and recall and is therefore not dominated by the majority non-hallucinated class.
Appendix~\ref{app:metric_choice} provides further discussion of alternative metrics

SHAM-Bench supports evaluation at two levels of granularity: token-level and sentence-level. Each sample in the dataset consists of a query $q$, a model-generated response $r$, and human-labeled hallucinated spans $S \subset r$. A detector predicts either hallucinated spans $\hat{S}$ or sentence-level labels $\hat{\mathbf{z}}$.
We convert all outputs to token-level binary labels. Given tokenization 
$\mathcal{T}(r) = (t_1, \dots, t_n)$, we define ground-truth labels 
$\mathbf{y} \in \{0,1\}^n$ and predicted labels $\hat{\mathbf{y}} \in \{0,1\}^n$ as:
\[
y_i = \mathbb{I}\!\left[\exists s \in S : t_i \mbox{ overlaps with } s\right].
\]
\[
\hat{y}_i = \mathbb{I}\!\left[\exists \hat{s} \in \hat{S} : t_i \mbox{ overlaps with } \hat{s}\right].
\]

For sentence-level evaluation, given segmentation $\mathcal{U}(r) = (u_1, \dots, u_m)$, we project span annotations onto sentences to obtain ground-truth labels $\mathbf{z} \in \{0,1\}^m$ and predicted labels $\hat{\mathbf{z}} \in \{0,1\}^m$:
\[
z_j = \mathbb{I}\!\left[\exists s \in S : u_j \text{ overlaps with } s\right]
\]
\[
\hat{z}_j = \mathbb{I}\!\left[\exists \hat{s} \in \hat{S} : u_j \text{ overlaps with } \hat{s}\right]
\]
Methods that directly output sentence labels use them as $\hat{\mathbf{z}}$ without projection. We report F$_1$ at both the span and sentence-level, aggregated by task and by dataset.
We additionally report precision and recall scores in Appendix \ref{app:full_results}.

\subsection{The SHAM-Bench Leaderboard}
\label{sec:leaderboard}
The SHAM-Bench leaderboard is publicly available at: \url{tbd}\footnote{Submissions must not include links to non-anonymous repositories. Will add in the camera-ready version.}, allowing research groups to submit their systems for automated evaluation.
As a \textit{living benchmark}, SHAM-Bench will evolve with new tasks and annotated datasets contributed by the community.
Its open-source dataset and evaluation code provide a unified platform for tracking progress in hallucination detection across diverse NLG tasks and domains.

\section{Hallucination Detection Methods}
\label{sec:methods}

Using SHAM-Bench, we evaluate seven existing high-performing hallucination detection methods in three major categories, namely discriminatively trained, generatively trained, and prompt-based methods. 
Appendix~\ref{app:method_details} provides detailed method descriptions.

\paragraph{Discriminatively trained methods.}
\label{subsec:disc}
\textbf{MiniCheck} \citep{tang-etal-2024-minicheck}, which is trained on synthetic data to judge whether claims are supported, and \textbf{AlignScore} \citep{zha-etal-2023-alignscore}, which is trained on multiple datasets to score alignment, both detect hallucination at the sentence level.
\textbf{PsiloQA} \citep{rykov-etal-2025-models}, trained on synthetic data, and \textbf{LettuceDetect} \citep{kovacs2025lettucedetect}, trained on human-annotated data, instead detect hallucination at the token level. These methods fundamentally treat hallucination detection as a binary classification problem whether at the sentence or token level.

\paragraph{Generatively trained methods.}
\label{subsec:gen}
\textbf{FAVA} \citep{mishra2024-fava} and \textbf{RAGTruth} \citep{niu-etal-2024-ragtruth} are train using the language modeling objective with  FAVA using synthetic data and RAGTruth using human annotated data. They both identify hallucinated content by generating the unsupported content as text spans.

\paragraph{Prompt-based methods.}
\label{subsec:prompt}
Prompt-based detectors use an LLM as a direct hallucination span extractor by providing the relevant information and instruction to the LLM as prompt. 
We evaluate both hand-crafted and automatically optimized prompts \citep{huang-etal-2025-ucsc}. 
Prompt templates are provided in Appendix~\ref{app:prompts}.

\begin{table*}[t]
  \centering
  \small
  \begin{tabular*}{\textwidth}{@{\extracolsep{\fill}} l@{\hspace{2em}}|ccccccc}
    \specialrule{0.9pt}{0pt}{3pt}
    \textbf{Method} & \textbf{Overall (std)} &
    \textbf{open\_qa} & \textbf{ctx\_qa} & \textbf{data2txt} &
    \textbf{open\_bio} & \textbf{summ} & \textbf{mt} \\
    \specialrule{0.7pt}{2pt}{2pt}
    \multicolumn{8}{l}{\hspace{2em}\textbf{Span-level}} \\
    \hline\addlinespace[2pt]
    PsiloQA (\texttt{ModernBERT-base}) & 0.239 (0.12) & 0.338 & 0.072 & 0.089 & \underline{0.363} & 0.359 & 0.216 \\
    LettuceDetect (\texttt{lettuce-large}) & 0.482 (0.21) & 0.324 & \underline{0.833} & \underline{0.614} & 0.198 & \textbf{0.541} & 0.382 \\
    RAGTruth (\texttt{ft-llama2-13B}) & \textbf{0.518} (0.19) & \underline{0.367} & \textbf{0.865} & \textbf{0.654} & 0.354 & 0.469 & 0.398 \\
    FAVA (\texttt{ft-llama2-7B}) & 0.381 (0.15) & 0.319 & 0.696 & 0.243 & 0.244 & 0.376 & \underline{0.411} \\
    Prompt (\texttt{GPT-5-mini}) & \underline{0.498} (0.07) & \textbf{0.402} & 0.603 & 0.528 & \textbf{0.525} & \underline{0.495} & \textbf{0.435} \\
    \specialrule{0.9pt}{2pt}{2pt}
    \multicolumn{8}{l}{\hspace{1.3em}\textbf{Sentence-level}} \\
    \hline\addlinespace[2pt]
    MiniCheck (\texttt{Bespoke-7B}) & 0.632 (0.12) & 0.603 & 0.466 & 0.492 & \underline{0.777} & \textbf{0.737} & 0.715 \\
    AlignScore (\texttt{roberta-large}) & 0.595 (0.16) & \underline{0.609} & 0.492 & 0.290 & 0.745 & 0.699 & 0.734 \\
    PsiloQA (\texttt{ModernBERT-base}) & 0.484 (0.24) & 0.572 & 0.094 & 0.214 & 0.754 & 0.599 & 0.671 \\
    LettuceDetect (\texttt{lettuce-large}) & \underline{0.638} (0.15) & 0.586 & \underline{0.842} & \underline{0.717} & 0.342 & 0.714 & 0.625 \\
    RAGTruth (\texttt{ft-llama2-13B}) & \underline{0.638} (0.13) & 0.579 & \textbf{0.869} & \textbf{0.752} & 0.533 & 0.590 & 0.507 \\
    FAVA (\texttt{ft-llama2-7B}) & 0.540 (0.16) & 0.539 & 0.701 & 0.268 & 0.435 & 0.572 & 0.728 \\
    Prompt (\texttt{GPT-5-mini}) & \textbf{0.691} (0.07) & \textbf{0.633} & 0.610 & 0.635 & \textbf{0.784} & \underline{0.726} & \textbf{0.759} \\
    \specialrule{0.9pt}{1pt}{1pt}
  \end{tabular*}
  \caption{Average span- and sentence-level F$_1$ scores on SHAM-Bench by task. Prompt method uses \texttt{gpt-5-mini} with reasoning but no prompt optimization. 
  \textbf{Overall} reports the unweighted mean and standard deviation of six task scores.
  Highest scores are in \textbf{bold}; second highest are \underline{underlined}.
  The full results can be found in Appendix~\ref{app:full_results}.}
  \label{tab:methods}
\end{table*}

\begin{table*}[t]
  \centering
  \small
  \setlength{\tabcolsep}{4pt}
  \begin{tabular*}{\textwidth}{@{\extracolsep{\fill}} l@{\hspace{2em}}|c c c c c c}
    \specialrule{0.9pt}{0pt}{3pt}
    \multirow{2}{*}{\textbf{Method}} &
    \multicolumn{3}{c}{\textbf{summ}} &
    \multicolumn{3}{c}{\textbf{qa}} \\
    \cmidrule(lr){2-4}\cmidrule(lr){5-7}
    &
    \textbf{faithbench} &
    \textbf{zhou} &
    \textbf{ragtruth} &
    \textbf{favabench} &
    \textbf{mushroom} &
    \textbf{ragtruth} \\
    \specialrule{0.7pt}{2pt}{2pt}
    \multicolumn{7}{l}{\hspace{2em}\textbf{Span-level}} \\
    \hline\addlinespace[2pt]
    PsiloQA (\texttt{ModernBERT-base}) & 0.225 & \underline{0.547} & 0.053 & 0.210 & \underline{0.556} & 0.072 \\
    LettuceDetect (\texttt{lettuce-large}) & 0.307 & \textbf{0.559} & 0.695 & 0.242 & 0.463 & \underline{0.833} \\
    RAGTruth (\texttt{ft-llama2-13B}) & \underline{0.323} & 0.357 & \textbf{0.835} & \textbf{0.327} & 0.435 & \textbf{0.865} \\
    FAVA (\texttt{ft-llama2-7B}) & 0.310 & 0.228 & \underline{0.761} & 0.235 & 0.463 & 0.695 \\
    Prompt (\texttt{GPT-5-mini}) & \textbf{0.387} & 0.514 & 0.541 & \underline{0.260} & \textbf{0.642} & 0.603 \\
    \specialrule{0.9pt}{2pt}{2pt}
    \multicolumn{7}{l}{\hspace{1.3em}\textbf{Sentence-level}} \\
    \hline\addlinespace[2pt]
    MiniCheck (\texttt{Bespoke-7B}) & 0.378 & 0.875 & 0.730 & 0.413 & \underline{0.926} & 0.466 \\
    AlignScore (\texttt{roberta-large}) & 0.348 & 0.864 & 0.626 & 0.428 & 0.915 & 0.492 \\
    PsiloQA (\texttt{ModernBERT-base}) & \underline{0.389} & \textbf{0.906} & 0.092 & 0.379 & 0.898 & 0.094 \\
    LettuceDetect (\texttt{lettuce-large}) & 0.363 & 0.845 & 0.713 & 0.410 & 0.885 & \underline{0.842} \\
    RAGTruth (\texttt{ft-llama2-13B}) & 0.349 & 0.566 & \textbf{0.845} & \underline{0.437} & 0.819 & \textbf{0.869} \\
    FAVA (\texttt{ft-llama2-7B}) & 0.334 & 0.570 & \underline{0.773} & 0.392 & 0.788 & 0.701 \\
    Prompt (\texttt{GPT-5-mini}) & \textbf{0.493} & \underline{0.885} & 0.569 & \textbf{0.450} & \textbf{0.944} & 0.610 \\
    \specialrule{0.9pt}{1pt}{1pt}
  \end{tabular*}
  \caption{Average span- and sentence-level F$_1$ scores on SHAM-Bench summarization and open/context\_qa tasks by dataset.
  Prompt method uses \texttt{gpt-5-mini} with reasoning but no prompt optimization.
  Highest scores are in \textbf{bold}; second highest are \underline{underlined}.}
  \label{tab:methods_by_dataset}
\end{table*}

\section{Benchmarking Results}
\label{sec:results}



\subsection{Prompt-based methods are robust across different tasks}
Results in Table~\ref{tab:methods} indicate prompted \texttt{GPT-5-mini} (with reasoning, no prompt optimization) is the most balanced method across all tasks.
At span-level F$_1$, the prompted method achieves an overall score of 0.498, placing second only to RAGTruth (0.518).
It ranks first on three of the six tasks: open\_qa (0.402), open\_bio (0.525), and mt (0.435), while falling behind RAGTruth or LettuceDetect on the three tasks that those models were directly trained on, which is further discussed in \S\ref{subsec:rq2}. 
At sentence level, it achieves the highest overall score (0.691), surpassing all discriminatively and generatively fine-tuned methods. It outperforms all discriminatively trained methods on four tasks, falling slightly behind MiniCheck on summarization and LettuceDetect on context\_qa. 
The prompt-based method also shows the most consistent performance across tasks, with the lowest standard deviation in task-level F$_1$ scores (0.07), suggesting that prompting is less prone to large task-specific performance swings.

\begin{table*}[t]
  \centering
  \small
  \begin{tabular*}{\textwidth}{@{\extracolsep{\fill}} l@{\hspace{-0.1em}}cc|ccccccc}
    \specialrule{0.9pt}{0pt}{3pt}
    \textbf{Method} & \textbf{Reason} & \textbf{Opt} & \textbf{Overall (std)} &
    \textbf{open\_qa} & \textbf{ctx\_qa} & \textbf{data2txt} &
    \textbf{open\_bio} & \textbf{summ} & \textbf{mt} \\
    \specialrule{0.7pt}{2pt}{2pt}
    \multicolumn{10}{l}{\hspace{2em}\textbf{Span-level}} \\
    \hline\addlinespace[2pt]
    prompt (\texttt{GPT-5-mini}) & \xmark & \xmark & 0.425 (0.09) & \underline{0.386} & 0.553 & 0.321 & \underline{0.486} & 0.471 & 0.331 \\
    prompt (\texttt{GPT-5-mini}) & \cmark & \xmark & \textbf{0.498} (0.07) & \textbf{0.402} & 0.603 & \underline{0.528} & \textbf{0.525} & \underline{0.495} & \underline{0.435} \\
    prompt (\texttt{GPT-5-mini}) & \cmark & \cmark & \underline{0.497} (0.09) & 0.364 & 0.636 & \textbf{0.536} & 0.476 & \textbf{0.532} & \textbf{0.437} \\
    \hline\addlinespace[2pt]
    prompt (\texttt{Qwen3-4B}) & \xmark & \xmark & 0.298 (0.07) & 0.343 & 0.176 & 0.255 & 0.362 & 0.306 & 0.347 \\
    prompt (\texttt{Qwen3-4B}) & \cmark & \xmark & 0.442 (0.11) & 0.369 & 0.647 & 0.511 & 0.328 & 0.457 & 0.338 \\
    prompt (\texttt{Qwen3-4B}) & \cmark & \cmark & 0.446 (0.15) & 0.329 & \underline{0.748} & 0.490 & 0.328 & 0.462 & 0.316 \\
    \hline\addlinespace[2pt]
    prompt (\texttt{Olmo3-7B-Inst}) & \xmark & \xmark & 0.383 (0.14) & 0.322 & 0.670 & 0.346 & 0.261 & 0.422 & 0.278 \\
    prompt (\texttt{Olmo3-7B-Think}) & \cmark & \xmark & 0.425 (0.07) & 0.355 & 0.539 & 0.495 & 0.384 & 0.442 & 0.338 \\
    prompt (\texttt{Olmo3-7B-Think}) & \cmark & \cmark & 0.435 (0.16) & 0.348 & \textbf{0.790} & 0.360 & 0.317 & 0.451 & 0.343 \\
    \specialrule{0.9pt}{1pt}{1pt}
  \end{tabular*}
  \caption{Average span-level F$_1$ scores of prompt-based detectors on SHAM-Bench by task. \textbf{Reason} and \textbf{opt} indicate whether reasoning and prompt optimization are used. \textbf{Overall} reports the unweighted mean and standard deviation of six task scores. Highest scores are in \textbf{bold}; second highest are \underline{underlined}. Complete span- and sentence-level results are in Table~\ref{tab:prompt_methods_full}.}
  \label{tab:prompt_methods}
\end{table*}

\label{subsec:rq1}


\subsection{Finetuned methods can perform well in narrow domains but struggle to generalize}

\label{subsec:rq2}

\paragraph{PsiloQA: strong where hallucinations are frequent, but over-predicts elsewhere.}
PsiloQA obtains its best span-F$_1$ on Mushroom QA (0.556) and Zhou summarization (0.547), two datasets with high hallucination rates, but it performs poorly on context\_qa (0.072) and data2txt (0.089), where hallucinations are rare.
The same pattern holds for sentence-level F$_1$: PsiloQA achieves its highest score  on Zhou summarization, where 91\% of responses are labeled as containing  hallucinations.
However, this over-prediction leads it to classify nearly all examples as hallucinated, dragging down its overall performance.


\paragraph{RAGTruth and LettuceDetect: dominant in-domain, weak elsewhere.}
Although RAGTruth achieves the highest overall span-F$_1$ (0.518), this aggregate masks a sharp split between its training tasks and everything else.
Table~\ref{tab:methods_by_dataset} makes this split explicit: RAGTruth scores 0.835 on the RAGTruth summarization partition and 0.865 on RAGTruth QA at the span level, but is only mediocre on other datasets.
The same pattern appears at sentence level, where it leads on its own partitions yet ranks among the worst on summarization Zhou and QA Mushroom.
LettuceDetect, fine-tuned on the same data, behaves similarly: strong on RAGTruth partitions but weak elsewhere, except on Zhou summarization, where over-flagging is less costly because most responses contain annotated hallucinations.

\paragraph{FAVA: surprisingly weak on its own dataset.}
Table~\ref{tab:methods_by_dataset} shows FAVA performs competitively on datasets where the evidence--response structure closely resembles FAVA's Wikipedia-grounded training format.  FAVA achieves span-F$_1$ of 0.761 on RAGTruth summarization and 0.695 on RAGTruth context\_qa.
However, it performs unexpectedly poorly on FAVABench (span-F$_1$ of 0.235), the benchmark derived from the same work that introduced FAVA.

\paragraph{MiniCheck and AlignScore: strong on text-only tasks, brittle on structured data.}
Table~\ref{tab:methods} shows the two sentence-level methods, MiniCheck and AlignScore, exhibit similar performance profiles despite the differences in architecture and training data. 
Similar to the PsiloQA, AlignScore is weak on data2txt, with a precision of 0.247 and recall of 0.429.
On this dataset it over-predicts hallucinated sentences when the input context contains structured JSON records, likely because it hasn't seen any JSON record grounding during claim-verification training, thus flagging response sentences as unsupported when JSON records are encountered.
MiniCheck performs better on data2txt likely because it is fine-tuned from a larger base model which brings broader pre-training coverage of structured data formats though still lags behind prompted method by a significant margin.

\subsection{Reasoning plays a significant role in prompt-based hallucination detection}
\label{subsec:prompt_compare}


\paragraph{Reasoning is the single most impactful factor.}
Enabling reasoning in prompt-based approaches improves accuracy across every model and task.
The gain is largest for \texttt{Qwen3-4B}, whose overall span-level F$_1$ jumps from 0.298 to 0.442 (+0.144) and sentence-level F$_1$ from 0.504 to 0.631 (+0.127).
\texttt{GPT-5-mini} improves more modestly (0.425 $\to$ 0.498 span-F$_1$), consistent with its stronger no-reasoning baseline; \texttt{OLMo-3-7B} similarly gains +0.042 span-F$_1$.

Reasoning substantially reduces false positives and false negatives.
On examples \emph{without} annotated hallucinations, \texttt{Qwen3-4B} with no reasoning averages only 0.13 span-F$_1$ because it often marks spans as hallucinated when they are not; reasoning raises this to 0.67.
\texttt{OLMo-3-7B-Instruct} without reasoning is instead over-conservative. It only scores 0.28 span-F$_1$ on all-hallucinated samples by rarely predicting any spans, leaving most hallucinations undetected. Reasoning raises it to 0.39.
Reasoning better balances over- and under-prediction, reducing both false positives and missed hallucinations.

\paragraph{Prompt optimization has mixed effects.}
MiPROv2 optimization improves performance on ctx\_qa and summarization, but it also consistently degrades open\_qa performance.
With only five examples per task used during optimization, the optimizer appears to over-adapt to these samples and generalize poorly.
This highlights the risk of prompt optimization with very small development sets.



\section{Challenges in Hallucination Detection}
\label{sec:challenges}

\begin{figure}[t]
  \centering
  \includegraphics[width=\linewidth]{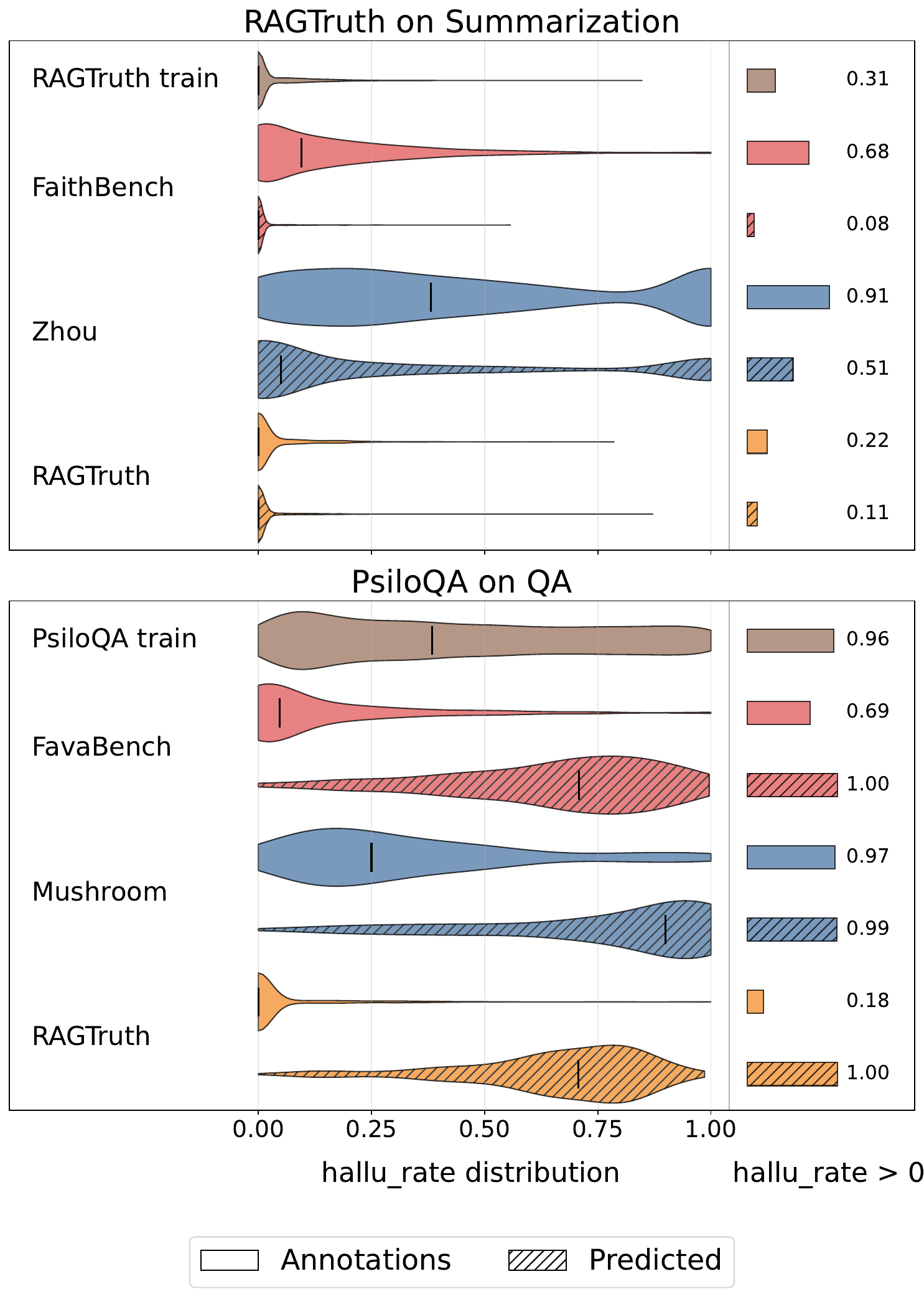}
  \caption{Distribution of hallucination rates as predicted by RAGTruth on summarization (top) and PsiloQA on QA (bottom). The first distribution in each plot is the distribution of the corresponding model's training set. To the right of each distribution is the proportion of examples that contain any hallucinated spans.}
  \label{fig:hallu_rate_violin}
  \end{figure}

To better understand the system behavior observed in \S\ref{sec:results}, we examine two complementary sources of difficulty. First, using RAGTruth and PsiloQA as representative cases, we analyze how the distribution of fine-tuning data shapes detector predictions. Second, we analyze failure cases from prompt-based methods and identify several factors that continue to limit span-level hallucination detection.


\subsection{Fine-tuned models tend to follow their training data distribution.}
As shown in Figure~\ref{fig:hallu_rate_violin}, hallucination rates in the RAGTruth summarization training set are concentrated near the low end. The model fine-tuned on this data closely follows the training distribution, regardless of the distribution of the evaluation set. This behavior of output distribution aligns well with the RAGTruth test set, but not with FaithBench, whose distribution is less skewed, or Zhou et al., which has an additional mode at the high end. As a result, the RAGTruth-trained model places disproportionate weight on low hallucination rates and tends to under-predict hallucinations.
The PsiloQA model exhibits the opposite bias. Although hallucination rates in the PsiloQA training set are more evenly distributed, 96\% of its training examples contain hallucinations. Consequently, when evaluated on QA datasets whose hallucination rates are concentrated near the low end, the model tends to over-predict hallucinations, resulting in systematically inflated hallucination estimates.

Fundamentally, these results reflect a train-test distribution shift, in which the training distribution does not match the evaluation setting. For practitioners, this suggests that class balancing alone is insufficient. Instead, training data should be curated to ensure adequate representation of hallucination cases, especially rare error types, while preserving or explicitly modeling the expected deployment distribution. 

\begin{figure}[t]
  \centering
  \includegraphics[width=\linewidth]{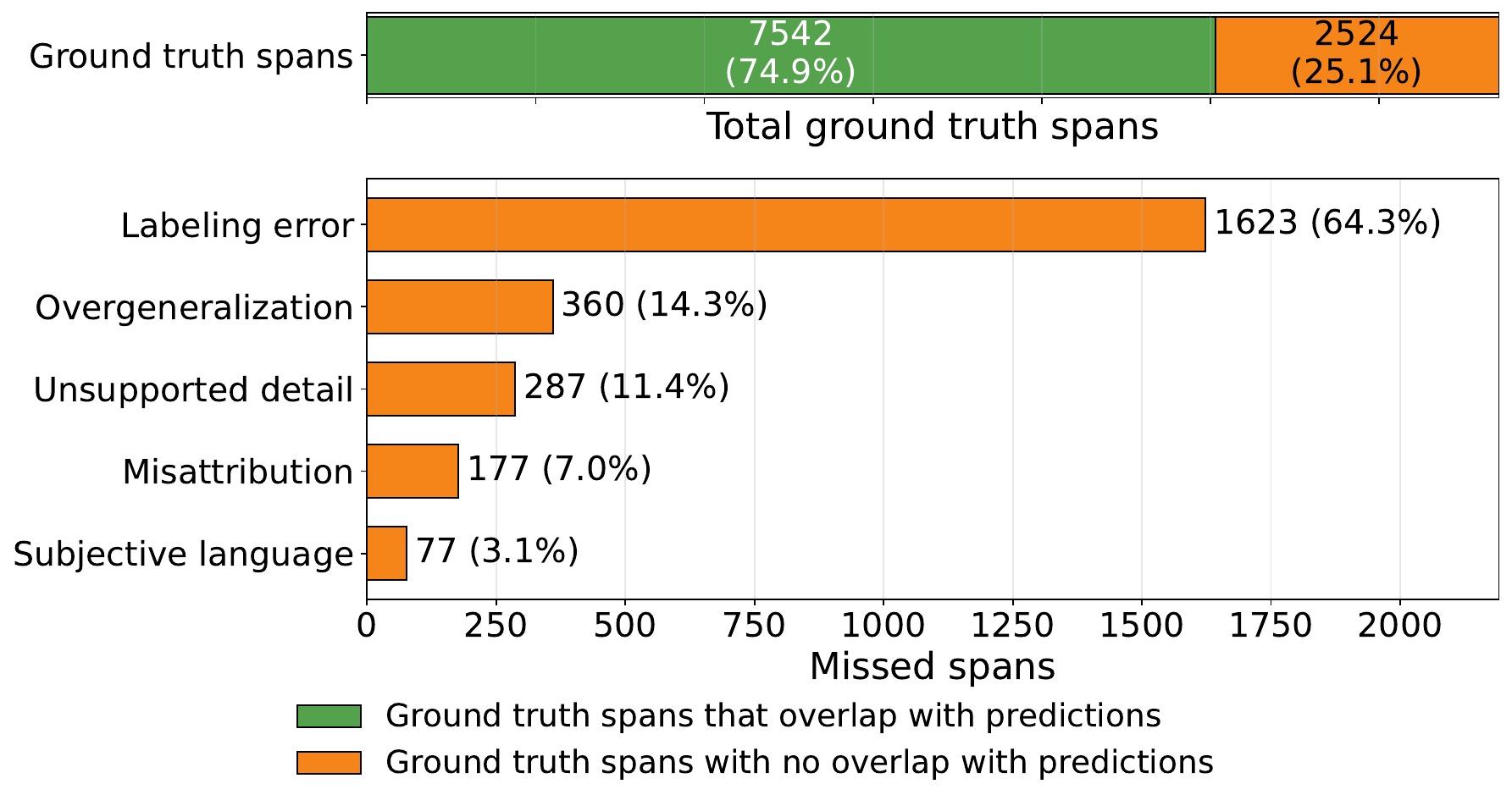}
  \caption{Top: ground-truth span recovery by GPT-5-mini reasoning prompt-based detector. Bottom: error categories for spans received no overlapping prediction.}
  \label{fig:groundtruth-miss}
  \end{figure}

\subsection{Prompt-based methods exhibit mixed failure modes}
To characterize prompt-based failure modes, we divide ground-truth spans into two groups: those that overlap with at least one prediction and those with no overlapping prediction. The first group captures cases in which the detector identifies the relevant factual issue, even if its predicted boundary differs from the human annotation. The second captures hallucinations that are missed entirely. For these missed spans, we use \texttt{GPT-5.4} to assign one of five error categories: \textit{labeling error}, \textit{overgeneralization}, \textit{unsupported detail}, \textit{misattribution}, and \textit{subjective language}; category definitions appear in Appendix~\ref{app:failure_categories}.

Figure~\ref{fig:groundtruth-miss} shows that the prompt-based detector recovers 7,542 of 10,066 ground-truth spans (74.9\%) with at least some overlap. This suggests that many apparent span-level errors arise from boundary disagreement rather than complete failure to recognize the underlying hallucination. Among overlapping spans, predicted spans are on average 23 words longer than the corresponding ground-truth span when they over-predict, and 31 words shorter when they under-predict. Over-prediction distribution shown in Appendix Figure~\ref{fig:overprediction-histograms}.

For the 2,524 ground-truth spans with no overlapping prediction, human labeling error accounts for the largest share (64\%). The remaining missed spans correspond to more substantive detector failures: overgeneralization accounts for 14\%, unsupported details for 11\%, misattribution for 7\%, and subjective language for 3\%. Except for misattribution, these categories generally involve subtle departures from the reference document, which remain difficult for prompt-based detectors to identify reliably.

These results suggest that improving span-level hallucination detection requires progress on annotation and modeling. Accurate annotations and standardized span definitions would reduce evaluation noise, while greater emphasis on subtle semantic departures from the reference could help detectors address the remaining missed hallucinations.


\section{Conclusion}
\label{sec:conclusion}

We presented SHAM-Bench, a unified benchmark for span-level hallucination detection across diverse generation tasks.
Our evaluation shows that fine-tuned detectors often perform well in-domain but degrade sharply outside their training distributions, whereas prompt-based methods with reasoning are more robust and consistent across tasks.
At the same time, ambiguous span boundaries, noisy labels, and subtle unsupported content remain open challenges for reliable span-level evaluation.
SHAM-Bench provides a foundation for future research on accurate, broadly generalizable hallucination detection.

\section{Limitations}
\label{sec:limitations}

\paragraph{Task coverage.}
SHAM-Bench covers a diverse set of NLG tasks, but it does not exhaustively represent all useful language generation settings.
Many important tasks are still missing because the benchmark is built from currently available human-annotated span-level hallucination resources.
Our goal is therefore not to provide a complete taxonomy of NLG tasks, but to establish a first extensible multi-task benchmark for span-level hallucination detection.
As more human-annotated span-level datasets become available, additional tasks and domains can be incorporated into SHAM-Bench.

\paragraph{RAGTruth over-representation.}
RAGTruth contributes 2,675 examples to SHAM-Bench across context QA, data-to-text, and part of the summarization task, accounting for approximately 41\% of the benchmark.
This over representation provides an advantage to detectors fine-tuned on the RAGTruth training set, meaning that strong overall performance does not necessarily reflect broad generalization.
As additional tasks and datasets are contributed to SHAM-Bench in the future this imbalance will diminish.


\paragraph{Legacy model outputs.}
A substantial portion of the Zhou data was generated by pre-LLM-era models.
The resulting errors differ significantly from hallucinations produced by modern LLMs.
This limits the extent to which performance on this subset reflects detector behavior in contemporary settings.

\paragraph{Annotation inconsistency.}
The source datasets used in SHAM-Bench were annotated by different groups following different annotation guidelines, which leads to substantial variation in how hallucination spans are defined and marked.
Some datasets favor minimal spans, while others use broader spans.
In aggregate, however, we believe that the rank of the hallucination detection methods will not be affected by variation in annotation styles across the individual the datasets.

\paragraph{Annotation quality.}
In this work, we did not perform quality control of the annotations provided in the source datasets. The quality of annotations across the datasets are known to vary widely, which limits the extent to which the benchmark can faithfully measure a model's ability to identify hallucinations.
However, again in aggregate, we believe that the relative ranking of hallucination detection methods is unlikely to be affected by the annotation quality of individual datasets.

\paragraph{Automated failure classification.}
The error categories used in our prompt-based failure analysis are assigned by \texttt{GPT-5.4} rather than manually verified by expert annotators.
As a result, the category proportions should be interpreted as an approximate diagnostic analysis rather than as definitive class labels.

\paragraph{Coverage of evaluated methods.}
In this work, we did not evaluate methods that require knowledge of or internal access to the generating model, such as hidden-state-based detectors or self-consistency methods.
To ensure faithful implementation and accurate evaluation, we encourage researchers developing such techniques to evaluate them on SHAM-Bench and submit their results to the SHAM-Bench Leaderboard.

\paragraph{Potential risks.}
Although SHAM-Bench is intended as a research benchmark, it may still introduce downstream risks if used incautiously.
Imperfect hallucination detectors can be over-trusted in deployment; false negatives may allow unsupported content to pass as reliable, while false positives may incorrectly flag factual content and reduce the usefulness of model outputs.
In addition, some benchmark instances involve generation about real people (e.g., the open\_biography task), so detector outputs may affect how biographical or factual claims are interpreted in downstream applications.
We therefore view SHAM-Bench as a research evaluation resource and not as a sufficient basis for high-stakes real-world deployment decisions.

\bibliography{custom}



\clearpage

\appendix

\section{Hallucination Detection Method Details}
\label{app:method_details}

\paragraph{Discriminatively trained methods.}
\textbf{MiniCheck} \citep{tang-etal-2024-minicheck} frames hallucination detection as \emph{evidence-grounded claim verification}: given a generated sentence and supporting context, it predicts whether the sentence is supported.
The models are trained with two synthetic data pipelines: Claim-to-Doc (C2D), which decomposes claims into atomic facts and builds multi-sentence evidence, and Doc-to-Claim (D2C), which derives summaries as claims from real documents and constructs unsupported pairs by removing sentences from the original document.
This yields sentence-level fact-checking models.

\textbf{AlignScore} \citep{zha-etal-2023-alignscore} frames hallucination detection as an \emph{information alignment} problem: for a text pair \((a,b)\), a unified alignment function \(f:(a,b)\rightarrow y\) estimates whether all information in \(b\) is present in, and non-contradictory with, \(a\).
The model is trained with multi-task learning on 15 datasets spanning seven NLP task families, and training is further augmented with synthetic positives from back translation and synthetic negatives from masking/infilling.
At inference time, it splits the output into sentence-level claims and chunks evidence into segments, computes the alignment score of each claim against all evidence chunks, and aggregates these scores into an overall score; sentences with low alignment are treated as likely hallucinated.

\textbf{PsiloQA} \citep{rykov-etal-2025-models} is an encoder token-level hallucination detector based on \texttt{ModernBERT} \citep{warner2024modernbert}.
It takes in context--question--response triples and predicts a binary classification label for each response token.
The model is trained on the English subset of the PsiloQA \citep{rykov-etal-2025-models} training data, which is synthetically generated from a three-stage pipeline: creating question-answer pairs from Wikipedia, eliciting potentially hallucinated answers from diverse LLMs in a no-context setting, and automatically annotating hallucinated spans by comparing responses to the golden answers and retrieved context using GPT-4o.

\textbf{LettuceDetect} \citep{kovacs2025lettucedetect} is a token-level hallucination detection model built on \texttt{ModernBERT} \citep{warner2024modernbert}.
Similar to PsiloQA, it processes concatenated context--question--response triples and predicts whether each response token is hallucinated.
In contrast to PsiloQA, LettuceDetect is trained using RAGTruth's dataset, which is labeled by human annotators.

\paragraph{Generatively trained methods.}
\textbf{FAVA} \citep{mishra2024-fava} frames hallucination detection as a generation problem.
Given an LLM response and retrieved evidence, the model generates the response with markers that wrap unsupported content.
FAVA is trained on synthetic data generated by first creating factually grounded seed passages from Wikipedia and then injecting errors while marking corrupted spans.
An LLM is trained on the resulting input--output pairs (erroneous input + evidence \(\rightarrow\) marked output).
Hallucinated spans are identified by marking them as unsupported content during generation.

\textbf{RAGTruth} \citep{niu-etal-2024-ragtruth} adopts a simpler generative approach by fine-tuning an LLM to directly output hallucinated text spans when given a context--response pair.
The LLM is fine-tuned on a large-scale training set (\(\sim\)18,000 examples) including context\_qa, data2text, and summarization tasks, which consists of human-annotated span-level labels of hallucinated content.

\paragraph{Prompt-based methods.}
We also evaluate prompt-based detection methods that treat an LLM as a direct hallucination span extractor.
Following \citet{huang-etal-2025-ucsc}, given a task question, evidence context, and generated response, the prompt asks the LLM to compare the response against the evidence and output hallucinated text as explicit spans, i.e. substrings of the response.
This approach requires no additional model training while still producing span-level outputs suitable for SHAM-Bench evaluation.
Beyond a hand-crafted instruction prompt, we apply prompt optimization \citep{huang-etal-2025-ucsc} using MiPROv2 \citep{opsahl-ong-etal-2024-optimizing} to maximize \(F_1\) scores.

\section{Metric Choice}
\label{app:metric_choice}

F$_1$ score is intentionally simple, but it is well aligned with the span-detection task evaluated in SHAM-Bench. Because gold annotations and model predictions are represented as sets of hallucinated response tokens, precision measures how many predicted hallucinated tokens are correct and recall measures how many gold hallucinated tokens are identified. F$_1$ combines these quantities into a single positive-class overlap score while avoiding the inflated performance that can arise from metrics dominated by the much larger set of non-hallucinated tokens.

Other overlap-based metrics would capture closely related information. For example, \citet{vazquez-etal-2025-semeval} uses intersection-over-union (IoU); for a fixed predicted token set and gold token set, IoU and F$_1$ are monotonic transformations of the same overlap counts, so they are expected to produce similar rankings. Threshold-independent metrics such as AUPRC and AUROC used in \citet{yeh-etal-2025-halluentity} can be useful when systems output calibrated confidence scores, but several SHAM-Bench systems, including prompt-based and generative span-extraction methods, output discrete spans without comparable scores. Balanced accuracy and macro-F$_1$ that provide balanced weighting of positive and negative classes reported in \citet{bao-etal-2025-faithbench} are also less directly aligned with our goal of measuring detection quality for the positive hallucinated-token class. We therefore use token-level F$_1$ as the primary span-level metric and report sentence-level F$_1$ as a complementary view when methods operate at sentence granularity.

\section{Failure Category Definitions}
\label{app:failure_categories}

We classify missed ground-truth spans into five categories for the prompt-based failure analysis in Section~\ref{sec:challenges}. These categories are intended to distinguish likely annotation artifacts from substantive detector failures, and to separate different types of unsupported content.

\paragraph{Labeling error.}
This category covers cases in which the missed ``ground-truth hallucination'' is not clearly a detector failure because the gold annotation itself is questionable. The annotated span may be directly supported by the source, may constitute an acceptable paraphrase or modest inference, or may reflect an annotation-boundary artifact, such as a fragment, duplicate span, discourse marker, or non-factual token. In such cases, the detector may have appropriately ignored the span or identified the substantive hallucination with a different boundary. This category therefore captures annotation-side noise rather than a clear failure to detect unsupported content.

\paragraph{Unsupported detail.}
This category covers genuine hallucination misses in which the response introduces concrete information that is not supported by the source. The unsupported content may be an added attribute, procedural step, recommendation, purpose, causal explanation, exact descriptor, or schema-derived claim. It also includes cases in which missing or null metadata is converted into an affirmative factual assertion, such as treating an unknown ambience field as evidence that a restaurant is ``casual'' or treating an unspecified amenity as absent. The defining property is that the response adds a specific fact-like detail that is not licensed by the available evidence.

\paragraph{Overgeneralization.}
Overgeneralization errors occur when the response remains loosely connected to the source but changes the strength, scope, certainty, or characterization of the supported claim. The response may turn narrow evidence into a broader rule, extrapolate from a single example to a general conclusion, or replace a source-supported description with a stronger or subtly different one. Unlike unsupported-detail errors, these cases often do not introduce a wholly fabricated fact; instead, they overextend or semantically drift from what the source establishes.

\paragraph{Misattribution.}
This category covers cases in which content may be related to the available evidence but is assigned to the wrong source, entity, task direction, comparison frame, status, or evidential basis. For example, a response may claim that a detail comes from a particular passage when it does not, merge information from separate cases, apply facts about one entity to another, or answer the reverse of the intended task. The central error is therefore not merely a lack of support, but the placement of otherwise plausible information in the wrong evidential or contextual frame.

\paragraph{Subjective language.}
This category covers unsupported evaluative, promotional, sentimental, or ambience-related wording. The response may add judgments such as ``friendly,'' ``cozy,'' ``popular,'' ``beautiful,'' or other claims about reception, service quality, or atmosphere that are not grounded in the source. Although these spans may sound plausible and are often softer than concrete factual fabrications, they still introduce unsupported characterizations. We separate this category because such language reflects a distinct form of embellishment rather than a specific factual addition.

\begin{figure}[t]
  \centering
  \includegraphics[width=\linewidth]{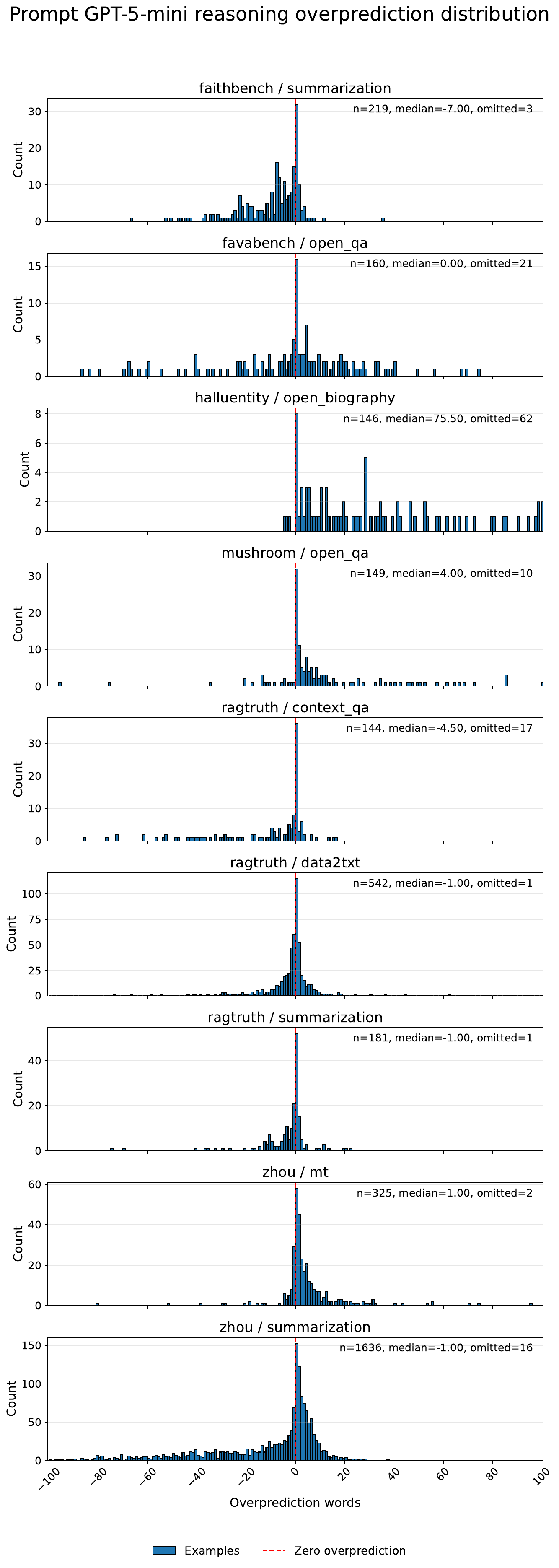}
  \caption{Distribution of over-prediction lengths for overlapping spans predicted by prompting GPT-5-mini with reasoning, grouped by dataset and query type. Negative values indicate under-prediction.}
  \label{fig:overprediction-histograms}
\end{figure}

\section{Full Experiment Results}
\label{app:full_results}
The complete prompt-based method results with sentence-level F$_1$ scores is given in Table~\ref{tab:prompt_methods_full}.
The full set of experimental results for span-level labeling with dataset breakdown can be found in Table~\ref{tab:span_by_dataset}, sentence-level results in Table~\ref{tab:sentence_by_dataset}, span-level precision/recall results in Table~\ref{tab:span_pr_by_dataset}, and sentence-level precision/recall results in Table~\ref{tab:sentence_pr_by_dataset}. 

\begin{table*}[t]
  \centering
  \small
  \begin{tabular*}{\textwidth}{@{\extracolsep{\fill}} l@{\hspace{-0.1em}}cc|ccccccc}
    \specialrule{0.9pt}{0pt}{3pt}
    \textbf{Method} & \textbf{Reason} & \textbf{Opt} & \textbf{Overall (std)} &
    \textbf{open\_qa} & \textbf{ctx\_qa} & \textbf{data2txt} &
    \textbf{open\_bio} & \textbf{summ} & \textbf{mt} \\
    \specialrule{0.7pt}{2pt}{2pt}
    \multicolumn{10}{l}{\hspace{2em}\textbf{Span-level}} \\
    \hline\addlinespace[2pt]
    prompt (\texttt{GPT-5-mini}) & \xmark & \xmark & 0.425 (0.09) & \underline{0.386} & 0.553 & 0.321 & \underline{0.486} & 0.471 & 0.331 \\
    prompt (\texttt{GPT-5-mini}) & \cmark & \xmark & \textbf{0.498} (0.07) & \textbf{0.402} & 0.603 & \underline{0.528} & \textbf{0.525} & \underline{0.495} & \underline{0.435} \\
    prompt (\texttt{GPT-5-mini}) & \cmark & \cmark & \underline{0.497} (0.09) & 0.364 & 0.636 & \textbf{0.536} & 0.476 & \textbf{0.532} & \textbf{0.437} \\
    \hline\addlinespace[2pt]
    prompt (\texttt{Qwen3-4B}) & \xmark & \xmark & 0.298 (0.07) & 0.343 & 0.176 & 0.255 & 0.362 & 0.306 & 0.347 \\
    prompt (\texttt{Qwen3-4B}) & \cmark & \xmark & 0.442 (0.11) & 0.369 & 0.647 & 0.511 & 0.328 & 0.457 & 0.338 \\
    prompt (\texttt{Qwen3-4B}) & \cmark & \cmark & 0.446 (0.15) & 0.329 & \underline{0.748} & 0.490 & 0.328 & 0.462 & 0.316 \\
    \hline\addlinespace[2pt]
    prompt (\texttt{Olmo3-7B-Inst}) & \xmark & \xmark & 0.383 (0.14) & 0.322 & 0.670 & 0.346 & 0.261 & 0.422 & 0.278 \\
    prompt (\texttt{Olmo3-7B-Think}) & \cmark & \xmark & 0.425 (0.07) & 0.355 & 0.539 & 0.495 & 0.384 & 0.442 & 0.338 \\
    prompt (\texttt{Olmo3-7B-Think}) & \cmark & \cmark & 0.435 (0.16) & 0.348 & \textbf{0.790} & 0.360 & 0.317 & 0.451 & 0.343 \\
    \specialrule{0.9pt}{2pt}{2pt}
    \multicolumn{10}{l}{\hspace{1.3em}\textbf{Sentence-level}} \\
    \hline\addlinespace[2pt]
    prompt (\texttt{GPT-5-mini}) & \xmark & \xmark & 0.641 (0.11) & 0.620 & 0.561 & 0.447 & 0.772 & 0.705 & 0.740 \\
    prompt (\texttt{GPT-5-mini}) & \cmark & \xmark & \underline{0.691} (0.07) & \textbf{0.633} & 0.610 & \underline{0.635} & \underline{0.784} & \underline{0.726} & \underline{0.759} \\
    prompt (\texttt{GPT-5-mini}) & \cmark & \cmark & \textbf{0.705} (0.06) & \underline{0.631} & 0.640 & \textbf{0.662} & \textbf{0.787} & \textbf{0.731} & \textbf{0.779} \\
    \hline\addlinespace[2pt]
    prompt (\texttt{Qwen3-4B}) & \xmark & \xmark & 0.504 (0.16) & 0.557 & 0.197 & 0.391 & 0.615 & 0.569 & 0.693 \\
    prompt (\texttt{Qwen3-4B}) & \cmark & \xmark & 0.631 (0.06) & 0.573 & 0.659 & 0.625 & 0.541 & 0.667 & 0.720 \\
    prompt (\texttt{Qwen3-4B}) & \cmark & \cmark & 0.630 (0.09) & 0.560 & \underline{0.758} & 0.591 & 0.524 & 0.610 & 0.736 \\
    \hline\addlinespace[2pt]
    prompt (\texttt{Olmo3-7B-Inst}) & \xmark & \xmark & 0.559 (0.12) & 0.530 & 0.675 & 0.354 & 0.484 & 0.612 & 0.697 \\
    prompt (\texttt{Olmo3-7B-Think}) & \cmark & \xmark & 0.620 (0.05) & 0.566 & 0.554 & 0.616 & 0.609 & 0.673 & 0.699 \\
    prompt (\texttt{Olmo3-7B-Think}) & \cmark & \cmark & 0.587 (0.14) & 0.553 & \textbf{0.790} & 0.362 & 0.500 & 0.614 & 0.702 \\
    \specialrule{0.9pt}{1pt}{1pt}
  \end{tabular*}
  \caption{Average span- and sentence-level F$_1$ scores of prompt-based hallucination detection methods on SHAM-Bench aggregated by task. \textbf{Reason} and \textbf{opt} indicate whether reasoning and prompt optimization are used. The \textbf{overall} column shows the unweighted mean and standard deviation of six task scores. Highest score in each column in \textbf{bold} and the second highest score is \underline{underlined}.}
  \label{tab:prompt_methods_full}
\end{table*}

\begin{sidewaystable}
  \centering
  \small
  \setlength{\tabcolsep}{4pt}
  \begin{tabular*}{\textwidth}{@{\extracolsep{\fill}} l c c | c c c c c c c c c @{}}
    \toprule
    \multirow{2}{*}{\textbf{Method}} &
    \multirow{2}{*}{\textbf{Reasoning}} &
    \multirow{2}{*}{\shortstack{\textbf{Prompt}\\\textbf{Optimization}}} &
    \multicolumn{1}{c}{\textbf{context\_qa}} &
    \multicolumn{1}{c}{\textbf{data2txt}} &
    \multicolumn{1}{c}{\textbf{mt}} &
    \multicolumn{1}{c}{\textbf{open\_bio}} &
    \multicolumn{2}{c}{\textbf{open\_qa}} &
    \multicolumn{3}{c}{\textbf{summ}} \\
    \cmidrule(lr){4-4}\cmidrule(lr){5-5}\cmidrule(lr){6-6}\cmidrule(lr){7-7}\cmidrule(lr){8-9}\cmidrule(lr){10-12}
    & & &
    \textbf{ragtruth} &
    \textbf{ragtruth} &
    \textbf{zhou} &
    \textbf{halluentity} &
    \textbf{favabench} &
    \textbf{mushroom} &
    \textbf{faithbench} &
    \textbf{ragtruth} &
    \textbf{zhou} \\
    \midrule
    PsiloQA (\texttt{ModernBERT-base}) & -- & -- & 0.072 & 0.089 & 0.216 & 0.363 & 0.210 & 0.556 & 0.225 & 0.053 & 0.547 \\
    LettuceDetect (\texttt{lettuce-large}) & -- & -- & \underline{0.833} & \underline{0.614} & 0.382 & 0.198 & 0.242 & 0.463 & 0.307 & 0.695 & \textbf{0.559} \\
    RAGTruth (\texttt{ft-llama2-13B}) & -- & -- & \textbf{0.865} & \textbf{0.654} & 0.398 & 0.354 & \textbf{0.327} & 0.435 & 0.323 & \textbf{0.835} & 0.357 \\
    FAVA (\texttt{ft-llama2-7B}) & -- & -- & 0.695 & 0.243 & 0.411 & 0.244 & 0.235 & 0.463 & 0.310 & 0.761 & 0.228 \\
    \hline\addlinespace[2pt]
    Prompt (\texttt{GPT-5-mini}) & \xmark & \xmark & 0.553 & 0.321 & 0.331 & \underline{0.486} & 0.261 & 0.597 & 0.353 & 0.497 & 0.504 \\
    Prompt (\texttt{GPT-5-mini}) & \cmark & \xmark & 0.603 & 0.528 & \underline{0.435} & \textbf{0.525} & 0.260 & \textbf{0.642} & \underline{0.387} & 0.541 & 0.514 \\
    Prompt (\texttt{GPT-5-mini}) & \cmark & \cmark & 0.636 & 0.536 & \textbf{0.437} & 0.476 & 0.271 & 0.520 & \textbf{0.422} & 0.575 & \underline{0.554} \\
    \hline\addlinespace[2pt]
    Prompt (\texttt{Qwen3-4B}) & \xmark & \xmark & 0.176 & 0.255 & 0.347 & 0.362 & 0.180 & \underline{0.620} & 0.178 & 0.176 & 0.413 \\
    Prompt (\texttt{Qwen3-4B}) & \cmark & \xmark & 0.647 & 0.511 & 0.338 & 0.328 & 0.234 & 0.597 & 0.333 & 0.720 & 0.385 \\
    Prompt (\texttt{Qwen3-4B}) & \cmark & \cmark & 0.748 & 0.490 & 0.316 & 0.328 & 0.262 & 0.442 & 0.353 & 0.753 & 0.372 \\
    \hline\addlinespace[2pt]
    Prompt (\texttt{Olmo-3-7B-Instruct}) & \xmark & \xmark & 0.670 & 0.346 & 0.277 & 0.261 & 0.199 & 0.532 & 0.251 & 0.592 & 0.409 \\
    Prompt (\texttt{Olmo-3-7B-Think}) & \cmark & \xmark & 0.538 & 0.495 & 0.338 & 0.384 & 0.214 & 0.595 & 0.327 & 0.594 & 0.416 \\
    Prompt (\texttt{Olmo-3-7B-Think}) & \cmark & \cmark & 0.790 & 0.360 & 0.343 & 0.317 & \underline{0.275} & 0.471 & 0.329 & 0.746 & 0.363 \\
    \bottomrule
  \end{tabular*}
  \caption{Average span-level F$_1$ of hallucination detection methods on SHAM-Bench grouped by task and dataset. Highest score in each column in \textbf{bold} and the second highest score is \underline{underlined}.}
  \label{tab:span_by_dataset}
\end{sidewaystable}

\begin{sidewaystable}
  \centering
  \small
  \setlength{\tabcolsep}{4pt}
  \begin{tabular*}{\textwidth}{@{\extracolsep{\fill}} l c c | c c c c c c c c c c @{}}
    \toprule
    \multirow{2}{*}{\textbf{Method}} &
    \multirow{2}{*}{\textbf{Reasoning}} &
    \multirow{2}{*}{\shortstack{\textbf{Prompt}\\\textbf{Optimization}}} &
    \multicolumn{1}{c}{\textbf{context\_qa}} &
    \multicolumn{1}{c}{\textbf{data2txt}} &
    \multicolumn{1}{c}{\textbf{mt}} &
    \multicolumn{1}{c}{\textbf{open\_bio}} &
    \multicolumn{2}{c}{\textbf{open\_qa}} &
    \multicolumn{3}{c}{\textbf{summ}} \\
    \cmidrule(lr){4-4}\cmidrule(lr){5-5}\cmidrule(lr){6-6}\cmidrule(lr){7-7}\cmidrule(lr){8-9}\cmidrule(lr){10-12}
    & & &
    \textbf{ragtruth} &
    \textbf{ragtruth} &
    \textbf{zhou} &
    \textbf{halluentity} &
    \textbf{favabench} &
    \textbf{mushroom} &
    \textbf{faithbench} &
    \textbf{ragtruth} &
    \textbf{zhou} \\
    \midrule
    MiniCheck (\texttt{Bespoke-7B}) & -- & -- & 0.466 & 0.493 & 0.715 & 0.777 & 0.413 & 0.926 & 0.378 & 0.730 & 0.875 \\
    AlignScore (\texttt{roberta-large}) & -- & -- & 0.492 & 0.290 & 0.734 & 0.745 & 0.428 & 0.915 & 0.348 & 0.626 & 0.864 \\
    PsiloQA (\texttt{ModernBERT-base}) & -- & -- & 0.094 & 0.214 & 0.671 & 0.754 & 0.379 & 0.898 & 0.389 & 0.092 & \textbf{0.906} \\
    LettuceDetect (\texttt{lettuce-large}) & -- & -- & \underline{0.842} & \underline{0.716} & 0.625 & 0.342 & 0.410 & 0.885 & 0.363 & 0.713 & 0.845 \\
    RAGTruth (\texttt{ft-llama2-13B}) & -- & -- & \textbf{0.869} & \textbf{0.752} & 0.507 & 0.533 & 0.437 & 0.819 & 0.349 & \textbf{0.845} & 0.566 \\
    FAVA (\texttt{ft-llama2-7B}) & -- & -- & 0.701 & 0.268 & 0.728 & 0.435 & 0.392 & 0.788 & 0.334 & 0.773 & 0.570 \\
    \hline\addlinespace[2pt]
    Prompt (\texttt{GPT-5-mini}) & \xmark & \xmark & 0.561 & 0.447 & 0.740 & 0.772 & 0.445 & 0.917 & 0.454 & 0.522 & 0.882 \\
    Prompt (\texttt{GPT-5-mini}) & \cmark & \xmark & 0.610 & 0.635 & \underline{0.759} & \underline{0.784} & \textbf{0.450} & \textbf{0.944} & \textbf{0.493} & 0.569 & \underline{0.885} \\
    Prompt (\texttt{GPT-5-mini}) & \cmark & \cmark & 0.640 & 0.662 & \textbf{0.779} & \textbf{0.787} & \underline{0.447} & \underline{0.943} & \underline{0.492} & 0.603 & 0.878 \\
    \hline\addlinespace[2pt]
    Prompt (\texttt{Qwen3-4B}) & \xmark & \xmark & 0.197 & 0.391 & 0.693 & 0.615 & 0.370 & 0.875 & 0.341 & 0.215 & 0.813 \\
    Prompt (\texttt{Qwen3-4B}) & \cmark & \xmark & 0.659 & 0.625 & 0.720 & 0.541 & 0.400 & 0.867 & 0.391 & 0.744 & 0.735 \\
    Prompt (\texttt{Qwen3-4B}) & \cmark & \cmark & 0.758 & 0.591 & 0.736 & 0.524 & 0.414 & 0.808 & 0.391 & \underline{0.774} & 0.618 \\
    \hline\addlinespace[2pt]
    Prompt (\texttt{Olmo-3-7B-Instruct}) & \xmark & \xmark & 0.675 & 0.354 & 0.697 & 0.484 & 0.354 & 0.827 & 0.334 & 0.605 & 0.719 \\
    Prompt (\texttt{Olmo-3-7B-Think}) & \cmark & \xmark & 0.554 & 0.616 & 0.699 & 0.609 & 0.402 & 0.843 & 0.424 & 0.618 & 0.790 \\
    Prompt (\texttt{Olmo-3-7B-Think}) & \cmark & \cmark & 0.790 & 0.361 & 0.701 & 0.500 & 0.414 & 0.788 & 0.382 & 0.757 & 0.636 \\
    \bottomrule
  \end{tabular*}
  \caption{Average sentence-level F$_1$ of hallucination detection methods on SHAM-Bench grouped by task and dataset. Highest score in each column is in \textbf{bold} and the second highest score is \underline{underlined}.}
  \label{tab:sentence_by_dataset}
\end{sidewaystable}

\begin{sidewaystable}
  \centering
  \small
  \setlength{\tabcolsep}{2pt}
  \begin{tabular*}{\textwidth}{@{\extracolsep{\fill}} l c c | c c c c c c c c c @{}}
    \toprule
    \multirow{2}{*}{\textbf{Method}} &
    \multirow{2}{*}{\textbf{Reasoning}} &
    \multirow{2}{*}{\shortstack{\textbf{Prompt}\\\textbf{Optimization}}} &
    \multicolumn{1}{c}{\textbf{context\_qa}} &
    \multicolumn{1}{c}{\textbf{data2txt}} &
    \multicolumn{1}{c}{\textbf{mt}} &
    \multicolumn{1}{c}{\textbf{open\_bio}} &
    \multicolumn{2}{c}{\textbf{open\_qa}} &
    \multicolumn{3}{c}{\textbf{summ}} \\
    \cmidrule(lr){4-4}\cmidrule(lr){5-5}\cmidrule(lr){6-6}\cmidrule(lr){7-7}\cmidrule(lr){8-9}\cmidrule(lr){10-12}
    & & &
    \textbf{ragtruth} &
    \textbf{ragtruth} &
    \textbf{zhou} &
    \textbf{halluentity} &
    \textbf{favabench} &
    \textbf{mushroom} &
    \textbf{faithbench} &
    \textbf{ragtruth} &
    \textbf{zhou} \\
    \midrule
    PsiloQA (\texttt{ModernBERT-base}) & -- & -- & 0.05 / 0.17 & 0.05 / 0.58 & 0.15 / 0.65 & 0.24 / 0.99 & 0.16 / 0.62 & 0.46 / 0.92 & 0.16 / 0.62 & 0.03 / 0.22 & 0.47 / 0.89 \\
    LettuceDetect (\texttt{lettuce-large}) & -- & -- & 0.84 / 0.84 & 0.63 / 0.67 & 0.36 / 0.53 & 0.16 / 0.31 & 0.22 / 0.45 & 0.37 / 0.88 & 0.35 / 0.31 & 0.70 / 0.72 & 0.52 / 0.75 \\
    RAGTruth (\texttt{ft-llama2-13B}) & -- & -- & 0.87 / 0.87 & 0.72 / 0.67 & 0.42 / 0.43 & 0.39 / 0.41 & 0.34 / 0.46 & 0.37 / 0.76 & 0.35 / 0.32 & 0.85 / 0.84 & 0.40 / 0.42 \\
    FAVA (\texttt{ft-llama2-7B}) & -- & -- & 0.71 / 0.69 & 0.26 / 0.25 & 0.43 / 0.47 & 0.46 / 0.23 & 0.27 / 0.35 & 0.49 / 0.63 & 0.34 / 0.31 & 0.77 / 0.76 & 0.40 / 0.22 \\
    \hline\addlinespace[2pt]
    Prompt (\texttt{GPT-5-mini}) & \xmark & \xmark & 0.55 / 0.58 & 0.27 / 0.58 & 0.26 / 0.66 & 0.42 / 0.67 & 0.26 / 0.41 & 0.51 / 0.90 & 0.40 / 0.37 & 0.49 / 0.55 & 0.50 / 0.65 \\
    Prompt (\texttt{GPT-5-mini}) & \cmark & \xmark & 0.61 / 0.62 & 0.51 / 0.64 & 0.38 / 0.66 & 0.49 / 0.64 & 0.26 / 0.42 & 0.55 / 0.94 & 0.51 / 0.36 & 0.55 / 0.56 & 0.55 / 0.61 \\
    Prompt (\texttt{GPT-5-mini}) & \cmark & \cmark & 0.63 / 0.66 & 0.50 / 0.71 & 0.37 / 0.70 & 0.40 / 0.71 & 0.23 / 0.51 & 0.41 / 0.96 & 0.51 / 0.40 & 0.57 / 0.62 & 0.51 / 0.76 \\
    \hline\addlinespace[2pt]
    Prompt (\texttt{Qwen3-4B}) & \xmark & \xmark & 0.21 / 0.18 & 0.32 / 0.28 & 0.31 / 0.55 & 0.58 / 0.32 & 0.28 / 0.22 & 0.75 / 0.63 & 0.32 / 0.16 & 0.20 / 0.18 & 0.51 / 0.46 \\
    Prompt (\texttt{Qwen3-4B}) & \cmark & \xmark & 0.66 / 0.65 & 0.56 / 0.55 & 0.28 / 0.59 & 0.45 / 0.31 & 0.31 / 0.28 & 0.65 / 0.72 & 0.40 / 0.32 & 0.73 / 0.73 & 0.47 / 0.42 \\
    Prompt (\texttt{Qwen3-4B}) & \cmark & \cmark & 0.76 / 0.75 & 0.51 / 0.53 & 0.25 / 0.63 & 0.36 / 0.36 & 0.30 / 0.34 & 0.40 / 0.76 & 0.39 / 0.35 & 0.76 / 0.77 & 0.37 / 0.49 \\
    \hline\addlinespace[2pt]
    Prompt (\texttt{Olmo-3-7B-Instruct}) & \xmark & \xmark & 0.67 / 0.68 & 0.35 / 0.36 & 0.22 / 0.57 & 0.31 / 0.39 & 0.21 / 0.33 & 0.52 / 0.73 & 0.28 / 0.29 & 0.59 / 0.60 & 0.42 / 0.55 \\
    Prompt (\texttt{Olmo-3-7B-Think}) & \cmark & \xmark & 0.56 / 0.54 & 0.57 / 0.51 & 0.29 / 0.54 & 0.53 / 0.36 & 0.29 / 0.25 & 0.64 / 0.71 & 0.43 / 0.32 & 0.61 / 0.59 & 0.51 / 0.46 \\
    Prompt (\texttt{Olmo-3-7B-Think}) & \cmark & \cmark & 0.79 / 0.79 & 0.36 / 0.36 & 0.29 / 0.60 & 0.39 / 0.32 & 0.33 / 0.32 & 0.44 / 0.72 & 0.40 / 0.32 & 0.75 / 0.75 & 0.40 / 0.43 \\
    \bottomrule
  \end{tabular*}
  \caption{Average span-level precision/recall (P/R) of hallucination detection methods on SHAM-Bench grouped by task and dataset. Each cell reports precision followed by recall, rounded to two decimal places.}
  \label{tab:span_pr_by_dataset}
\end{sidewaystable}

\begin{sidewaystable}
  \centering
  \small
  \setlength{\tabcolsep}{2pt}
  \begin{tabular*}{\textwidth}{@{\extracolsep{\fill}} l c c | c c c c c c c c c @{}}
    \toprule
    \multirow{2}{*}{\textbf{Method}} &
    \multirow{2}{*}{\textbf{Reasoning}} &
    \multirow{2}{*}{\shortstack{\textbf{Prompt}\\\textbf{Optimization}}} &
    \multicolumn{1}{c}{\textbf{context\_qa}} &
    \multicolumn{1}{c}{\textbf{data2txt}} &
    \multicolumn{1}{c}{\textbf{mt}} &
    \multicolumn{1}{c}{\textbf{open\_bio}} &
    \multicolumn{2}{c}{\textbf{open\_qa}} &
    \multicolumn{3}{c}{\textbf{summ}} \\
    \cmidrule(lr){4-4}\cmidrule(lr){5-5}\cmidrule(lr){6-6}\cmidrule(lr){7-7}\cmidrule(lr){8-9}\cmidrule(lr){10-12}
    & & &
    \textbf{ragtruth} &
    \textbf{ragtruth} &
    \textbf{zhou} &
    \textbf{halluentity} &
    \textbf{favabench} &
    \textbf{mushroom} &
    \textbf{faithbench} &
    \textbf{ragtruth} &
    \textbf{zhou} \\
    \midrule
    MiniCheck (\texttt{Bespoke-7B}) & -- & -- & 0.45 / 0.52 & 0.43 / 0.67 & 0.71 / 0.73 & 0.84 / 0.77 & 0.35 / 0.64 & 0.92 / 0.96 & 0.42 / 0.37 & 0.72 / 0.75 & 0.88 / 0.88 \\
    AlignScore (\texttt{roberta-large}) & -- & -- & 0.48 / 0.53 & 0.25 / 0.43 & 0.74 / 0.74 & 0.79 / 0.76 & 0.36 / 0.67 & 0.91 / 0.94 & 0.38 / 0.34 & 0.62 / 0.64 & 0.86 / 0.86 \\
    PsiloQA (\texttt{ModernBERT-base}) & -- & -- & 0.07 / 0.18 & 0.13 / 0.64 & 0.66 / 0.69 & 0.65 / 1.00 & 0.29 / 0.71 & 0.88 / 0.95 & 0.30 / 0.68 & 0.06 / 0.23 & 0.91 / 0.91 \\
    LettuceDetect (\texttt{lettuce-large}) & -- & -- & 0.84 / 0.85 & 0.70 / 0.77 & 0.63 / 0.62 & 0.37 / 0.34 & 0.36 / 0.59 & 0.89 / 0.91 & 0.39 / 0.36 & 0.71 / 0.72 & 0.84 / 0.84 \\
    RAGTruth (\texttt{ft-llama2-13B}) & -- & -- & 0.88 / 0.87 & 0.78 / 0.75 & 0.51 / 0.51 & 0.78 / 0.45 & 0.45 / 0.52 & 0.85 / 0.81 & 0.36 / 0.34 & 0.85 / 0.84 & 0.57 / 0.57 \\
    FAVA (\texttt{ft-llama2-7B}) & -- & -- & 0.70 / 0.70 & 0.28 / 0.27 & 0.73 / 0.73 & 0.72 / 0.34 & 0.40 / 0.49 & 0.85 / 0.76 & 0.35 / 0.33 & 0.78 / 0.77 & 0.57 / 0.57 \\
    \hline\addlinespace[2pt]
    Prompt (\texttt{GPT-5-mini}) & \xmark & \xmark & 0.56 / 0.59 & 0.37 / 0.67 & 0.73 / 0.75 & 0.82 / 0.77 & 0.42 / 0.57 & 0.92 / 0.94 & 0.46 / 0.48 & 0.50 / 0.57 & 0.88 / 0.88 \\
    Prompt (\texttt{GPT-5-mini}) & \cmark & \xmark & 0.61 / 0.63 & 0.58 / 0.77 & 0.75 / 0.77 & 0.84 / 0.78 & 0.41 / 0.61 & 0.94 / 0.97 & 0.55 / 0.47 & 0.55 / 0.61 & 0.88 / 0.88 \\
    Prompt (\texttt{GPT-5-mini}) & \cmark & \cmark & 0.64 / 0.66 & 0.62 / 0.79 & 0.78 / 0.79 & 0.86 / 0.77 & 0.39 / 0.63 & 0.94 / 0.96 & 0.55 / 0.47 & 0.59 / 0.63 & 0.88 / 0.88 \\
    \hline\addlinespace[2pt]
    Prompt (\texttt{Qwen3-4B}) & \xmark & \xmark & 0.21 / 0.20 & 0.42 / 0.42 & 0.69 / 0.70 & 0.81 / 0.53 & 0.44 / 0.41 & 0.94 / 0.85 & 0.41 / 0.33 & 0.21 / 0.23 & 0.81 / 0.81 \\
    Prompt (\texttt{Qwen3-4B}) & \cmark & \xmark & 0.67 / 0.66 & 0.65 / 0.64 & 0.72 / 0.73 & 0.81 / 0.45 & 0.48 / 0.42 & 0.94 / 0.84 & 0.43 / 0.38 & 0.75 / 0.75 & 0.74 / 0.73 \\
    Prompt (\texttt{Qwen3-4B}) & \cmark & \cmark & 0.77 / 0.76 & 0.62 / 0.59 & 0.73 / 0.75 & 0.78 / 0.43 & 0.47 / 0.46 & 0.88 / 0.79 & 0.43 / 0.37 & 0.78 / 0.78 & 0.62 / 0.62 \\
    \hline\addlinespace[2pt]
    Prompt (\texttt{Olmo-3-7B-Instruct}) & \xmark & \xmark & 0.68 / 0.68 & 0.35 / 0.36 & 0.69 / 0.71 & 0.62 / 0.47 & 0.35 / 0.45 & 0.88 / 0.83 & 0.35 / 0.35 & 0.60 / 0.61 & 0.72 / 0.72 \\
    Prompt (\texttt{Olmo-3-7B-Think}) & \cmark & \xmark & 0.56 / 0.56 & 0.62 / 0.65 & 0.69 / 0.71 & 0.80 / 0.53 & 0.45 / 0.44 & 0.93 / 0.81 & 0.47 / 0.41 & 0.62 / 0.62 & 0.79 / 0.79 \\
    Prompt (\texttt{Olmo-3-7B-Think}) & \cmark & \cmark & 0.79 / 0.79 & 0.36 / 0.36 & 0.70 / 0.71 & 0.77 / 0.40 & 0.48 / 0.44 & 0.87 / 0.76 & 0.42 / 0.37 & 0.76 / 0.76 & 0.64 / 0.64 \\
    \bottomrule
  \end{tabular*}
  \caption{Average sentence-level precision/recall (P/R) of hallucination detection methods on SHAM-Bench grouped by task and dataset. Each cell reports precision followed by recall, rounded to two decimal places.}
  \label{tab:sentence_pr_by_dataset}
\end{sidewaystable}




\clearpage
\onecolumn
\raggedbottom
\section{Example Labeled Outputs}
\label{app:examples}

\begin{figure}[H]
\begin{center}
\fbox{%
\begin{minipage}{0.97\textwidth}
\small
\raggedright
\textbf{Question}

Summarize the following text

\medskip

\textbf{Context}

Lake Providence is a town in and the parish seat of East Carroll Parish in northeastern Louisiana, United States. University of Mississippi. About 55 percent of its undergraduates and 60 percent overall come from Mississippi, and 23 percent are minorities; international students come from 90 nations.

\medskip

\textbf{Response with gold annotated spans}

\fbox{%
\begin{minipage}{0.97\textwidth}
\small
\raggedright
Lake Providence is a town in Louisiana that is the parish seat of East Carroll Parish. The University of Mississippi has a student body that is \colorbox{green!25}{55\% undergraduate} and 60\% overall from Mississippi. 23\% of the student body are minorities and international students come from 90 nations.
\end{minipage}%
}

\medskip

\textbf{Response with predicted spans}

\fbox{%
\begin{minipage}{0.97\textwidth}
\small
\raggedright
Lake Providence is a town in Louisiana that is the parish seat of East Carroll Parish. The University of Mississippi \colorbox{red!20}{has a student body that is 55\% undergraduate} and 60\% overall from Mississippi. 23\% of the student body are minorities and international students come from 90 nations.
\end{minipage}%
}

\medskip

\textbf{Metrics}

Span: F1=0.5000, Precision=0.3333, Recall=1.0000\\
Sentence: F1=1.0000, Precision=1.0000, Recall=1.0000
\end{minipage}%
}
\end{center}
\caption{Example FaithBench summarization output labeled by prompted \texttt{GPT-5-mini} with reasoning. The box shows the task question, grounding context, model response, gold hallucinated spans, model predicted spans, and the resulting metrics.}
\end{figure}

\begin{figure}[H]
\begin{center}
\fbox{%
\begin{minipage}{0.97\textwidth}
\small
\raggedright
\textbf{Question}

Translate the following text from Chinese to English:

\begin{CJK*}{UTF8}{gbsn}
附连插口10具有两个保持段11和12（图6显示了它们的截面），两个升离斜坡13，以及在保持段11和升离斜坡13之间的接合处的旋转止动部14。
\end{CJK*}

\medskip

\textbf{Context}

attachment socket 10 has two retaining segments 11 and 12 whose cross-section is shown in fig. 6, two lift-off ramps 13 and a rotational stop 14 at the junction between retaining segment 11 and a lift-off ramp 13.

\medskip

\textbf{Response with gold annotated spans}

\fbox{%
\begin{minipage}{0.97\textwidth}
\small
\raggedright
\begin{CJK*}{UTF8}{gbsn}
the attached plug 10 has two maintaining sections 11 and 12 \colorbox{green!25}{(图}) 6 shows their intersections, two uphill slopes 13, and a rotating stop unit 14 in the joints between the maintaining sections 11 and uphill slopes 13.
\end{CJK*}
\end{minipage}%
}

\medskip

\textbf{Response with predicted spans}

\fbox{%
\begin{minipage}{0.97\textwidth}
\small
\raggedright
\begin{CJK*}{UTF8}{gbsn}
the \colorbox{red!20}{attached plug 10} has two maintaining sections 11 and 12 (\colorbox{red!20}{图 6 shows their intersections}), \colorbox{red!20}{two uphill slopes 13}, and a rotating stop unit 14 \colorbox{red!20}{in the joints between the maintaining sections 11 and uphill slopes 13.}
\end{CJK*}
\end{minipage}%
}

\medskip

\textbf{Metrics}

Span: F1=0.0769, Precision=0.0417, Recall=0.5000\\
Sentence: F1=1.0000, Precision=1.0000, Recall=1.0000
\end{minipage}%
}
\end{center}
\caption{Example Zhou machine-translation output labeled by prompted \texttt{GPT-5-mini} with reasoning. The box shows the task question, grounding context, model response, gold hallucinated spans, model predicted spans, and the resulting metrics.}
\end{figure}

\begin{figure}[H]
\begin{center}
\fbox{%
\begin{minipage}{0.97\textwidth}
\small
\raggedright
\textbf{Question}

how to factory reset a htc one vx

\medskip

\textbf{Context}

passage 1: 1 In order to switch off the device, hold down the Power button for a while. 2 Then press and hold the Volume Down + Power key for a couple of seconds. 3 When you see Bootloader Mode menu release held buttons. 4 From the menu choose option RECOVERY MODE by using Volume buttons to scroll and Power key to select options. 4 Congratulations! 5 You should be right now in the expected mode.

\medskip

passage 2: 1 First of all, switch off the phone by using the Power rocker. 2 After that, press and hold Volume Down and Power buttons together for a couple of seconds. 3 You can release all keys when the Bootloader Mode appears on the screen. 4 Choose the Factory Reset by pressing the Volume Down button to scroll and Power key to select.

\medskip

passage 3: Be sure to back up any data and files you want to keep before you do a factory reset. Hard reset your HTC One VX as follows: 1 Slide the Notifications panel open. 2 Tap Storage. 3 Tap Factory data reset. 4 Tap Reset phone, and then tap Erase everything.

\medskip

\textbf{Response with gold annotated spans}

\fbox{%
\begin{minipage}{0.97\textwidth}
\small
\raggedright
Based on the provided passages, here are the steps to factory reset an HTC One VX:

\medskip

1. Slide the Notifications panel open.\\
2. Tap on Storage.\\
3. Tap on Factory data reset.\\
4. Tap on Reset phone.\\
5. Tap on Erase everything.

\medskip

Alternatively, you can also perform a factory reset by following these steps:

\medskip

1. Hold down the Power button until the device shuts down.\\
2. Press and hold the Volume Down and Power buttons together for a few seconds until the bootloader mode appears.\\
3. Use the Volume down button to scroll to ``Recovery Mode'' and press the Power button to select.\\
4. From the recovery menu, select ``\colorbox{green!25}{wipe data}/factory reset'' and press the Power button to confirm.\\
5. Wait for the process to complete.\\
\colorbox{green!25}{\parbox[t]{0.9\linewidth}{6. Once the reset is done, select ``Reboot System'' from the recovery menu and press the Power button to restart the device.}}

\medskip

Please note that performing a factory reset will erase all data on the device, so make sure to back up any important data before doing so.
\end{minipage}%
}

\medskip

\textbf{Response with predicted spans}

\fbox{%
\begin{minipage}{0.97\textwidth}
\small
\raggedright
Based on the provided passages, here are the steps to factory reset an HTC One VX:

\medskip

1. Slide the Notifications panel open.\\
2. Tap on Storage.\\
3. Tap on Factory data reset.\\
4. Tap on Reset phone.\\
5. Tap on Erase everything.

\medskip

Alternatively, you can also perform a factory reset by following these steps:

\medskip

1. Hold down the Power button until the device shuts down.\\
2. Press and hold the Volume Down and Power buttons together for a few seconds until the bootloader mode appears.\\
3. Use the Volume down button to scroll to ``Recovery Mode'' and press the Power button to select.\\
4. \colorbox{red!20}{From the recovery menu, select ``wipe Data/Factory Reset'' and press the Power button to confirm.}\\
5. \colorbox{red!20}{Wait for the process to complete.}\\
6. \colorbox{red!20}{\parbox[t]{0.84\linewidth}{Once the reset is done, select ``Reboot System'' from the recovery menu and press the Power button to restart the device.}}
\medskip

Please note that performing a factory reset will erase all data on the device, so make sure to back up any important data before doing so.
\end{minipage}%
}

\medskip

\textbf{Metrics}

Span: F1=0.6750, Precision=0.5294, Recall=0.9310\\
Sentence: F1=0.6667, Precision=0.6667, Recall=0.6667
\end{minipage}%
}
\end{center}
\caption{Example RAGTruth context\_qa output labeled by prompted \texttt{GPT-5-mini} with reasoning. The box shows the task question, grounding passages, model response, gold hallucinated spans, model predicted spans, and the resulting metrics.}
\end{figure}

\section{Prompt Templates}
\label{app:prompts}
\textbf{System prompt}
\begin{CJK*}{UTF8}{gbsn}
\begin{Verbatim}[breaklines=true,breakanywhere=true,breaksymbolleft={},breaksymbolright={}]
Based on the provided information, identify incorrect spans in the given answer text.

You will be provided with a question with some related information. You will also be given an answer to the question. Your task is to identify any specific parts of the answer that describes facts that are not supported by the provided information. If there are multiple incorrect segments, report each one separately.

# Steps
1. **Read the Question and Information**: Carefully read the provided question and information.
2. **Analyze the Answer**: Carefully evaluate the given answer for accuracy regarding the question and the provided information.
3. **Identify Incorrect Spans**: Mark the parts of the text that are incorrect, misleading, or irrelevant to the question and context provided in the question.

# Output Format
The output should be in JSON format as shown below:
```json
{
  "incorrect_spans": [
    "identified incorrect span",
    "another identified incorrect span"
  ]
}
```
If no incorrect spans are identified, return an empty list:
```json
{
  "incorrect_spans": []
}
```

# Example 1
**Input**:
<question>
What is the capital of France?

Answer the question based on the following information:
Paris, the capital city of France, is a metropolis steeped in history, culture, and global significance. This comprehensive analysis will delve into the city's current status, basic information, and historical importance, providing a thorough understanding of why Paris is not just the capital of France, but also one of the world's most influential cities.
</question>

<answer>
The capital of France is Berlin.
</answer>

**Output**:
```json
{
  "incorrect_spans": [
    "Berlin"
  ]
}
```

# Example 2
**Input**:
<question>
Answer the question: "What did Mara do?" based on the following paragraph:

The train arrived at the platform without a sound, its windows glowing softly as if lit by bottled moonlight. Mara hesitated before stepping aboard, aware that everyone else seemed to know exactly where they were going, while she alone carried a ticket with no destination printed on it. As the doors closed, the city outside blurred into a ribbon of color and memory, and she realized with a quiet thrill that whatever waited at the end of the line would not be something she could ever return from--only something she could become.
</question>

<answer>
Mara got off a train that arrived without a sound while holding a ticket to Copenhagen.
</answer>

**Output**:
```json
{
  "incorrect_spans": [
    "got off",
    "Copenhagen"
  ]
}
```

# Example 3
**Input**:
<question>
Translate the following text from Chinese to English:
请去帮我买点零食。

The reference translation is:
Please go and buy me buy some snacks.
</question>

<answer>
Please help me buy groceries.
</answer>

**Output**:
```json
{
  "incorrect_spans": [
    "groceries"
  ]
}
```

# Example 4
**Input**:
<question>
Summarize the following news:
The Riverton City Council voted unanimously Tuesday night to approve a six-month pilot program that will introduce late-night bus service across five major routes. The program, scheduled to begin in March, is intended to support shift workers, expand transportation access, and reduce nighttime traffic incidents linked to fatigue and impaired driving.

Under the plan, buses will run every 30 minutes between midnight and 4 a.m., connecting downtown Riverton with several residential neighborhoods and the industrial district on the city's east side. Transportation Director Lena Morales said the routes were selected based on employment data, hospital locations, and feedback from residents collected during a public survey last fall.

Funding for the pilot will come from a combination of state transit grants and unused funds from a postponed road expansion project. City officials emphasized that ridership levels, operating costs, and safety outcomes will be closely monitored throughout the trial period. At the end of six months, the council will vote on whether to make the service permanent, expand it, or discontinue the program altogether.
</question>

<answer>
The Riverton City Council approved a six-month pilot program to run late-night buses on five major routes starting in March. The service will operate every 30 minutes from midnight to 4 a.m., targeting shift workers and improving nighttime safety. Officials will monitor costs, ridership, and safety before deciding whether to continue or expand the program.
</answer>

**Output**:
```json
{
  "incorrect_spans": []
}
```

# Notes
- It is possible for multiple incorrect spans to exist in the same answer; make sure to capture each one.
- If the answer is fully correct, return `"incorrect_spans": []`.
- Try to identify short spans as short as possible that captures the incorrect part of the answer.
- The spans should appear in the same order as they appear in the original answer.
\end{Verbatim}
\end{CJK*}

\noindent\textbf{User prompt}

\begin{Verbatim}[breaklines=true,breakanywhere=true,breaksymbolleft={},breaksymbolright={}]
<question>
{query}
</question>

<answer>
{answer}
</answer>
\end{Verbatim}




\end{document}